\documentclass[journal,compsoc]{IEEEtran}
\usepackage[breaklinks=true,colorlinks,citecolor=black,linkcolor=black,urlcolor=black]{hyperref}
\usepackage[dvipsnames,table,xcdraw]{xcolor}

\usepackage[utf8]{inputenc} 
\usepackage[T1]{fontenc}    
\usepackage{url}            
\usepackage{booktabs}       
\usepackage{amsfonts}       
\usepackage{nicefrac}       
\usepackage{microtype}      
\usepackage{graphicx}
\usepackage{wrapfig}
\usepackage{makecell}
\usepackage{amsmath}
\usepackage{float}
\usepackage{amssymb}
\usepackage{bm}
\usepackage{overpic}
\usepackage{multicol}
\usepackage{multirow}
\usepackage{siunitx}
\usepackage{enumitem}
\usepackage{colortbl}
\usepackage{pifont}
\usepackage{bbm}

\newcommand{\cmark}{\ding{51}}%

\newcommand{\figref}[1]{Fig.~\ref{#1}}
\newcommand{\tabref}[1]{Tab.~\ref{#1}}

\newcommand{\secref}[1]{Sec.~\ref{#1}}
\newcommand{\myPara}[1]{\vspace{6pt}\noindent\textbf{#1}}
\newcommand{\sArt}{state-of-the-art~}

\def\cc{\cellcolor[gray]{.95}}

\definecolor{wrq_color}{RGB}{16, 172, 132}

\newcommand{\revised}[1]{\textcolor{black}{#1}} 

\newcommand{\AP}[1]{AP$_{#1}$}
\newcommand{\conv}[1]{${#1}\times{#1}$}

\def\methodname{YOLO-MS}
\def\protocolname{HKS}
\def\blockname{MS-Block}
\def\fullblockname{MS-Block}
\def\attentionname{GQL}
\def\sota{SOTA}

\def\ie{\emph{i.e.\ }}
\def\eg{\emph{e.g., }}

\def\down{$\downarrow$}
\graphicspath{{./figures/}}

\begin{document}

\title{\methodname{}: Rethinking Multi-Scale Representation Learning for Real-time Object Detection}

\author{Yuming Chen, Xinbin Yuan, Jiabao Wang, Ruiqi Wu, Xiang Li,
Qibin Hou,~\IEEEmembership{Member,~IEEE}, \\
and Ming-Ming Cheng,~\IEEEmembership{Senior Member,~IEEE}
\IEEEcompsocitemizethanks{
\IEEEcompsocthanksitem All authors are with VCIP, School of Computer Science, Nankai University, Tianjin, China (Corresponding author: Qibin Hou).
\IEEEcompsocthanksitem This research was supported by NSFC (NO. 62225604, No. 62276145), the Fundamental Research Funds for the Central Universities (Nankai University, 070-63223049). Computations were supported by the Supercomputing Center of Nankai University (NKSC).
}
}

\markboth{IEEE TRANSACTIONS ON PATTERN ANALYSIS AND MACHINE INTELLIGENCE}%
{Shell \MakeLowercase{\textit{et al.}}: A Sample Article Using IEEEtran.cls for IEEE Journals}


\IEEEtitleabstractindextext{%
\begin{abstract}
We aim at providing the object detection community with an efficient and performant object detector, termed \methodname{}.
The core design is based on a series of investigations on how multi-branch features of the basic block and convolutions with different kernel sizes affect the detection performance of objects at different scales.
The outcome is a new strategy that can significantly enhance multi-scale feature representations of real-time object detectors.
To verify the effectiveness of our work, we train our \methodname{} on the MS COCO dataset from scratch without relying on any other large-scale datasets, like ImageNet or pre-trained weights.
Without bells and whistles, our \methodname{} outperforms the recent state-of-the-art real-time object detectors, including YOLO-v7, RTMDet, and YOLO-v8.
Taking the XS version of \methodname{} as an example, it can achieve an AP score of 42+\% on MS COCO, which is about 2\% higher than RTMDet with the same model size.
Furthermore, our work can also serve as a plug-and-play module for other YOLO models.
Typically, our method significantly advances the AP{s}, AP{l}, and AP of YOLOv8-N from 18\%+, 52\%+, and 37\%+ to 20\%+, 55\%+, and 40\%+, respectively, with even fewer parameters and MACs.
Code and trained models are publicly available at \url{https://github.com/FishAndWasabi/YOLO-MS}. \\
We also provide the Jittor version at \url{https://github.com/NK-JittorCV/nk-yolo}.
\end{abstract}

}

\maketitle

\section{Introduction}
\label{sec:intro}

\revised{Real-time object detectors, exemplified by the YOLO series \cite{redmon2016yolo, redmon2016yolov2, redmon2018yolov3, bochkovskiy2020yolov4, wang2021scaledyolov4, jocher2020yolov5, ge2021yolox, li2022yolov6, wang2022yolov7,lyu2022rtmdet, xu2022pp, jocher2023yolov8, lv2023rtdetr, wang2023goldyolo, wang2024yolov9, wang2024yolov10}, have found significant applications in industries, particularly for edge devices, such as drones and robotics.}
Different from previous object detectors \cite{zhu2021deformable,zhang2023dino,ren2015faster,carion2020end,tian2019fcos, dai2016r}, real-time object detectors aim to pursue an optimal trade-off between speed and accuracy.
To this end, numerous works have been proposed to develop efficient and powerful architectures for real-time object detection \cite{redmon2018yolov3, wang2020cspnet, wang2022designing, wang2022yolov7}.
From the first generation of DarkNet \cite{redmon2018yolov3} to CSPNet \cite{wang2020cspnet}, and then to the recent extended ELAN \cite{wang2022yolov7}, the architectures of real-time object detectors have experienced significant changes accompanied by rapid performance improvements.

\begin{figure}[ht]
  \footnotesize
  \centering
  \begin{overpic}[width=0.95\linewidth]{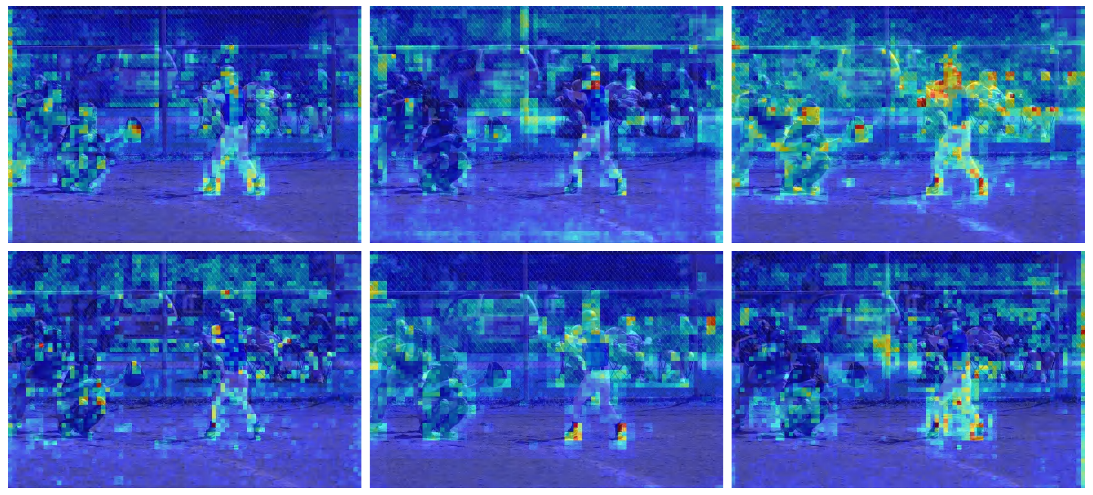}
    \put(10, -3){ELAN \cite{lyu2022rtmdet}}
    \put(44, -3){CSP \cite{wang2022yolov7}}
    \put(74, -3){Res2Net \cite{gao2021res2net}}
    \put(-3, 2){\rotatebox{90}{Last Branch}}
    \put(-3, 25){\rotatebox{90}{First Branch}}
  \end{overpic} \\
  \vspace{20pt}
  \setlength{\tabcolsep}{15pt}
  \begin{tabular}{l|cccccc}
    Model   & YOLOv7 & RTMDet & Res2Net  \\ \hline
    Params  & 6.2M   & 4.9M   & 4.9M     \\
    MACs    & 6.9G   & 8.1G   & 8.1G     \\
    AP{s}   & 19.9   & 20.7   & 21.3     \\
    $\mathcal{D}$  & 1.6$\times$1e-3  & 2.1$\times$1e-3  & 2.9$\times$1e-3\\ 
  \end{tabular}
  \vspace{-5pt}
  \caption{Branch feature diversity for different YOLO models. 
    For simplicity, only the feature visualization results of the two branches are presented, and it suffices to show the effectiveness of our method in enriching feature diversity.
    In the table, $\mathcal{D}$ is an intuitive indicator to measure the diversity of detectors' inter-branch features\protect\footnotemark[1].
  }\label{fig:branch_feature_diversity}
\end{figure}

\begin{figure*}[t]
  \centering
  \begin{overpic}[width=0.48\textwidth]{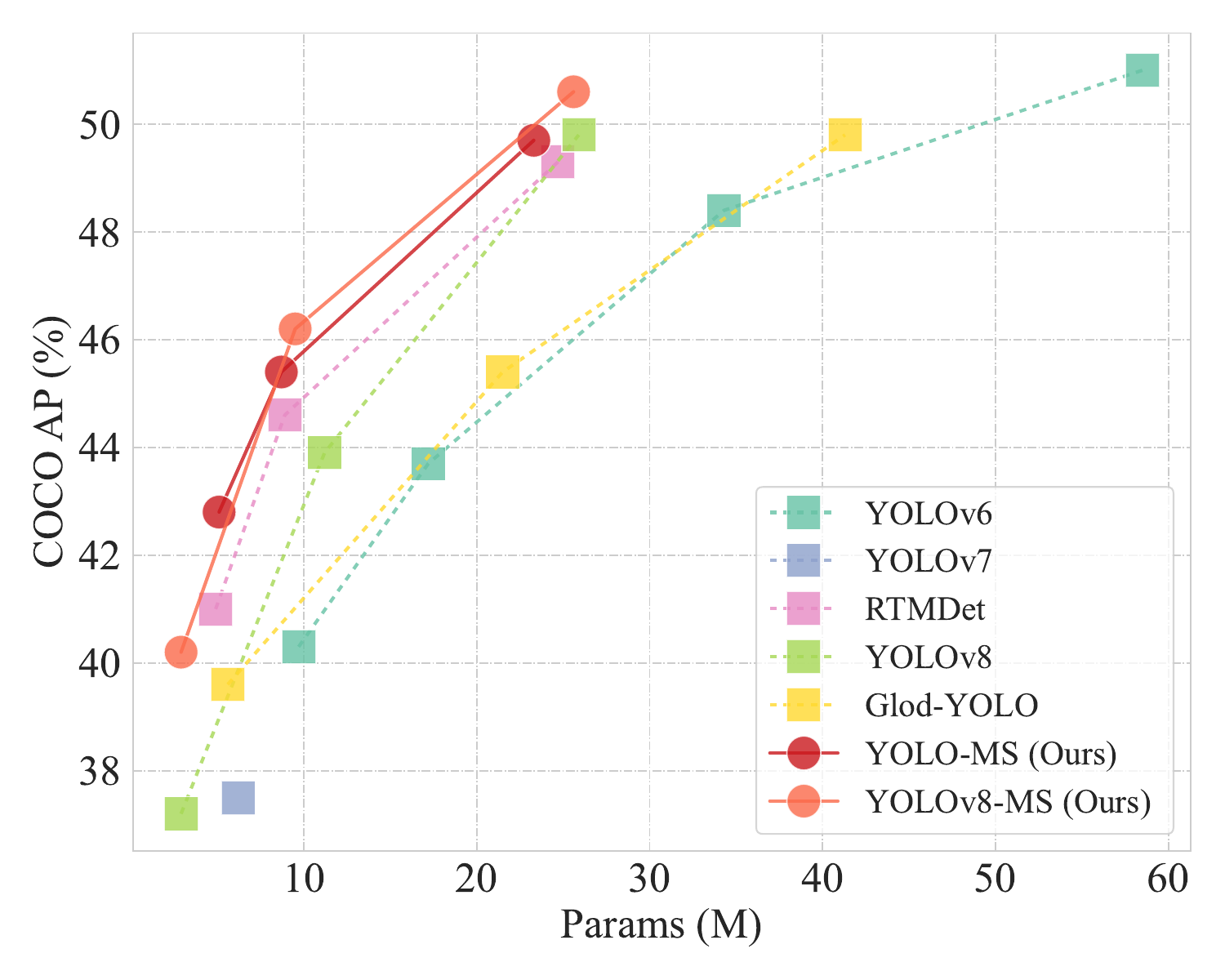}
    \put(80, 37.2){~~\scriptsize\cite{li2022yolov6}}
    \put(80, 33.2){~~\scriptsize\cite{wang2022yolov7}}
    \put(80, 29.2){~~\scriptsize\cite{lyu2022rtmdet}}
    \put(80, 25.2){~~\scriptsize\cite{jocher2023yolov8}}
    \put(83, 21.4){~~\scriptsize\cite{wang2023goldyolo}}
    \put(50, -1){(a)}
  \end{overpic} \hfill
  \begin{overpic}[width=0.48\textwidth]{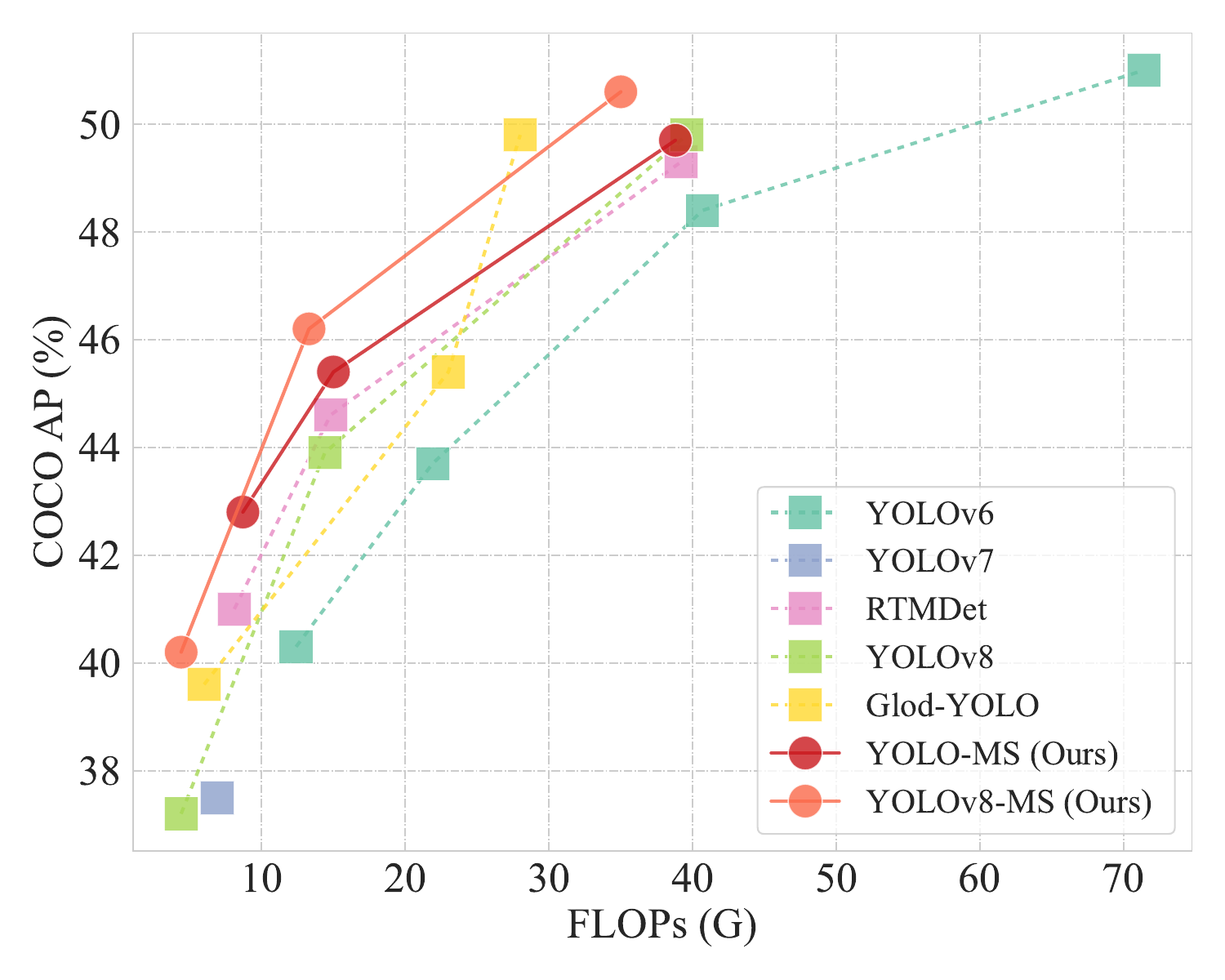}
    \put(80, 37.2){~~\scriptsize\cite{li2022yolov6}}
    \put(80, 33.2){~~\scriptsize\cite{wang2022yolov7}}
    \put(80, 29.2){~~\scriptsize\cite{lyu2022rtmdet}}
    \put(80, 25.2){~~\scriptsize\cite{jocher2023yolov8}}
    \put(83, 21.4){~~\scriptsize\cite{wang2023goldyolo}}
    \put(50, -1){(b)}
  \end{overpic}
  \vspace{-5pt}
  \caption{
    Comparisons with other \sArt real-time object detectors on the MS COCO dataset \cite{lin2014microsoft}.
    (a) \AP{} performance v.s. \#parameters.
    (b) \AP{} performance v.s. \#computations (MACs).
    The input size utilized to compute the MACs is $640 \times 640$.  
    Our proposed \methodname{} achieves the best trade-off between performance and computations.
}
\label{fig:performance_curve}
\end{figure*}

Albeit these works have achieved impressive performance, recognizing objects at different scales remains a fundamental challenge for real-time object detectors.
Existing real-time object detectors prioritize leveraging improvements on the macro-structure to augment the multi-scale feature representations.
Typically, FPN \cite{lin2017feature}, PAFPN \cite{liu2018pafpn}, and Gather-Distribute mechanism \cite{wang2023goldyolo} modify the model neck to improve the aggregation of different scale features.
Despite the success, these methods overlook the importance of learning multi-scale feature representations in the basic building blocks. 
Recent advancements in the YOLO series \cite{wang2022yolov7,wang2020cspnet} introduce multi-branch structures, which can be considered a method for implicitly learning multi-scale features.
\footnotetext[1]{
Suppose $N_{d}$ images are in the given dataset and $N_{m}$ blocks in the given model.
Let ${\mathbf{f}_{i}(d,m)}$ denote feature of the $i^{th}$ branch in the $m^{th}$ block for the $d^{th}$ image,
where $\mathbf{f} \in R^L$ and $L$ is the dimension of $\mathbf{f}$. 
$\mathcal{D} = \frac{1}{N_{m}N_{d}N_{b}}\sum_{d=1}^{N_{d}}\sum_{m=1}^{N_{m}}\sum_{i=1}^{N_{b}-1}\sum_{j=i+1}^{N_{b}} \frac{\|\mathbf{f}_{i}(d,m) - \mathbf{f}_{j}(d,m)\|_2}{L},$ 
where $\|\cdot\|_2$ denotes the Euclidean Norm and $N_{b}$ is branch number of the basic block.
Higher $\mathcal{D}$ represents greater diversity among inter-branch features.
}
However, visualizations and statistical analyses presented in \figref{fig:branch_feature_diversity} show that their inter-branch features are homogenized, reflecting a potential absence of diverse multi-scale information.
Another solution proposed to the above issue is the Res2Net \cite{gao2021res2net} block structure.
Although incorporating the hierarchical multi-branch structure of Res2Net can enrich the diversity of inter-branch features compared to the blocks above, it introduces contaminative spatial information as shown in \figref{fig:branch_feature_diversity}, not related to the target object, potentially harming detection performance.

\begin{figure*}[t]
  \centering
  \small
  \begin{overpic}[width=0.9\textwidth]{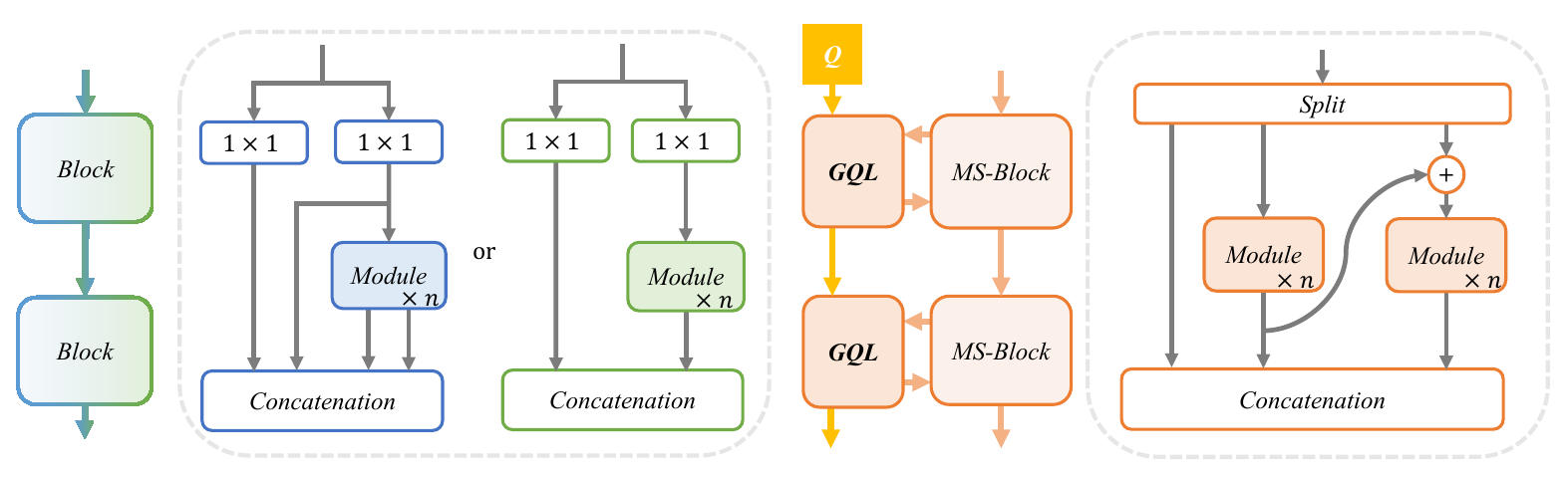}
    \put(14,-1){ELAN Series \cite{wang2022yolov7}}
    \put(34,-1){CSP Series \cite{wang2020cspnet}}
    \put(81,-1){\fullblockname{}}
    \put(30,-3){(a)}
    \put(82,-3){(b)}
  \end{overpic}
  \setlength{\abovecaptionskip}{15pt}
  \vspace{-2pt}
  \caption{
    (a) Architecture of fundamental blocks widely used in the previous YOLO models, \eg ELAN \cite{wang2020cspnet} series or CSP \cite{wang2020cspnet} series.
    (b) Architecture of proposed \fullblockname{}.
    $n$ refers to the number of modules (we use the inverted bottleneck module \cite{sandler2018mobilenetv2}).
    $Q$ denotes the global query used in the \attentionname{}.
  }\label{fig:detailed_structure_compare}
  \vspace{-8pt}
\end{figure*}

In this paper, we rethink how to encode multi-scale feature representations in multi-branch building blocks.
We argue that, for each building block, besides encoding multi-scale features as done in \cite{gao2021res2net}, it is also crucial to dynamically aggregate different granularities of information from different branches.
The detection head at each feature level in FPN \cite{lin2017feature} is responsible for detecting objects of sizes spanning over a certain range, \eg from 32 to 64 for the short side of the bounding box.
We design a \textbf{G}lobal \textbf{Q}uery \textbf{L}earning (GQL) strategy and present a new type of building block named \fullblockname{}.
As illustrated in \figref{fig:detailed_structure_compare}, we maintain a global query that stores the cross-stage spatial representations of each branch.
The GQL enables each block to dynamically balance the influence of each branch according to input and stage location.

Moreover, instead of using homogeneous kernel sizes for all the blocks, as done in previous real-time object detectors, we propose a \textbf{H}eterogeneous \textbf{K}ernel Size \textbf{S}election (HKS) protocol in different \fullblockname{s}
that gradually increases the convolutions' kernel size as the network goes deeper.
Specifically, we use small-kernel convolutions in shallow layers to process the high-resolution features more efficiently while adopting large-kernel convolutions in deep layers to capture high-level semantic information for better recognizing large objects.

Based on the above design principles, we present our real-time object detector, termed \textbf{\methodname{}}.
To evaluate the performance of our \methodname{}, we conduct comprehensive experiments on the MS COCO \cite{lin2014microsoft} dataset.
Quantitative comparisons with other \sArt methods are also provided to showcase the strong performance of our method.
As demonstrated in~\figref{fig:performance_curve}, \methodname{} outperforms other recent real-time object detectors with a better computation-performance trade-off.
To be specific, our \methodname{}-XS achieves an AP score of 42.8\% on MS COCO, with only 5.1M learnable parameters and 8.7G MACs.
Additionally, \methodname{}-S and \methodname{} yield 45.4\% AP and 52.1\% AP, respectively, with 8.7M and 50.8 learnable parameters, 
surpassing the baseline RTMDet \cite{lyu2022rtmdet}.
Besides, our work can also be used as a plug-and-play module to bring improvement to other YOLO models.
Typically, our method advances the AP of YOLOv8-n from 37\%+ to 40\%+ with even fewer parameters and MACs.

\section{Related Work}
\label{sec:related_work}

\subsection{Real-Time Object Detection}
 
The task of object detection is to detect objects in a specific scene.
Despite the outstanding performance achieved by the multi-stage detectors \cite{girshick2014rcnn, girshick2015fast, ren2015faster, lin2017feature, he2017mask,  cai2018cascade, wang2019region, pang2019lrcnn, chen2019hybrid} and end-to-end detectors \cite{zhu2021deformable, carion2020end, meng2021conditional, li2022dn, liu2022dabdetr, zhang2023dino}, their complicated structures often hinder them from achieving real-time performance, which is a prerequisite for real-world applications for object detection.
Much effort has gone into developing efficient detectors in network architecture and training techniques to achieve the optimal trade-off between speed and accuracy.
As opposed to the previous two-stage detectors, most real-time object detection networks adopt the one-stage framework \cite{lin2017focal, tian2019fcos, zheng2020distance, zhang2020bridging, li2020generalized, li2021gfocalv2}.
In particular, the YOLO series is the most prominent representative.

The architecture design is the main concern during the development of YOLO because it is a key factor in model performance.
Starting from the ancestor of the YOLO family, \ie, YOLOv1 \cite{redmon2016yolo}, the network architectures have undergone dramatic changes.
YOLOv4 \cite{bochkovskiy2020yolov4, wang2021scaledyolov4} modifies the DarkNet with cross-stage partial connections (CSPNet) \cite{wang2020cspnet} for better performance, and many YOLOs are proposed based on it, \eg YOLOv5 \cite{jocher2020yolov5}, RTMDet \cite{lyu2022rtmdet}, and YOLOv8 \cite{jocher2023yolov8}.
To be specific, RTMDet \cite{lyu2022rtmdet} first introduces large-kernel convolutions ($5\times 5$) into networks to boost the feature extraction capability of the CSP block.
YOLOv6 \cite{li2022yolov6} and PPYOLOE \cite{xu2022pp} explore the re-parameterization techniques for higher accuracy without additional inference costs in YOLO.
YOLOv7 \cite{wang2022yolov7} proposes Extended Efficient Layer Aggregation Networks (E-ELAN) that can learn and converge effectively by controlling the shortest, longest gradient path.
RT-DETR \cite{lv2023rtdetr} constructs the first transformer-based model to avoid the negative impact of NMS at inference time.
Unlike the above methods, which introduce new training or optimization techniques, this paper focuses on improving real-time object detectors by learning more expressive multi-scale feature representations.
%

\subsection{Multi-Scale Feature Representation Learning}
Multi-Scale feature representation learning in computer vision has been studied for a long time \cite{witkin1984scale, chen2017rethinking, zhang2021rstnet, gao2021rfnext, yin2024camoformer,guo2022segnext,hou2024conv2former}.
A strong multi-scale feature representation capability can effectively boost the model performance, which has been demonstrated in many tasks \cite{lin2017feature, liu2018pafpn, he2015spp, zhang2025referring, xu2022pp}, including real-time object detection \cite{redmon2018yolov3,bochkovskiy2020yolov4, jocher2020yolov5, li2022yolov6, lyu2022rtmdet}.

\myPara{Multi-scale feature learning in real-time object detection.}
Many real-time object detectors extract multi-scale features by integrating features from different feature levels in the neck part \cite{lin2017feature,liu2018pafpn}.
For instance, YOLOv3 \cite{redmon2018yolov3} and later YOLO series introduce FPN \cite{lin2017feature}, PAFPN \cite{liu2018pafpn}, and Gather-Distribute mechanism \cite{wang2023goldyolo}, respectively, to enhance the fusion of multi-scale features.
The SPP \cite{he2015spp} module is widely utilized to enlarge the receptive field of the encoder. 
Additionally, multi-scale data augmentations \cite{redmon2018yolov3} are widely used as effective training skills to improve multi-scale ability.
However, the mainstream of the basic building block overlooks the importance of multi-scale feature representation learning while focusing on how to facilitate detection efficiency or how to introduce new training techniques, especially for the CSP \cite{wang2020cspnet} block and the ELAN \cite{wang2022yolov7, wang2022designing} block.
In contrast, our method concentrates on the design of the basic block and analyzes how the inter-branch features affect the ability to represent multi-scale features.

\myPara{Large-kernel convolutions.}
Our work is also related to using convolutions with different kernel sizes, especially large-kernel convolutions.
Recently, large-kernel convolutions have been revitalized depth-wise \cite{ding2022scaling}.
The wider receptive field provided by large-kernel convolutions can be a potent technique for building strong multi-scale feature representations \cite{guo2022visual, guo2022segnext,hou2024conv2former,liu2022convnet,li2022sge, li2023large}.
In the real-time object detection field, RTMDet \cite{lyu2022rtmdet} first attempts to adopt large-kernel convolutions into networks.
However, the kernel size only reaches $5\times 5$ due to speed limitation.
Homogeneous block design in different stages limits the application of large-kernel convolutions.
This paper proposes a \protocolname{} protocol based on some empirical findings.
Our \protocolname{} protocol adopts convolutional layers with different kernel sizes at different stages, enabling a good trade-off between speed and accuracy with the help of large-kernel convolutions.

\section{Methodology}

As a crucial aspect of modern object detectors, learning multi-scale feature representations significantly influences detection performance \cite{lin2017feature,liu2018pafpn}.
In this section, aiming to construct a real-time object detector with a robust multi-scale ability, we propose \fullblockname{} and \protocolname{} protocol.
Our \fullblockname{} comprises an enhanced hierarchical multi-branch structure to advance the diversity of inter-branch features, along with \attentionname{} to provide cross-stage guidance for reducing harmful spatial information and enhancing multi-scale representations.
The \protocolname{} protocol is introduced to incorporate large-kernel convolution efficiently and effectively in real-time object detectors, aiming for a better multi-scale ability.

\begin{figure*}[t]
    \centering
    \footnotesize
    \begin{overpic}
        [width=\textwidth]{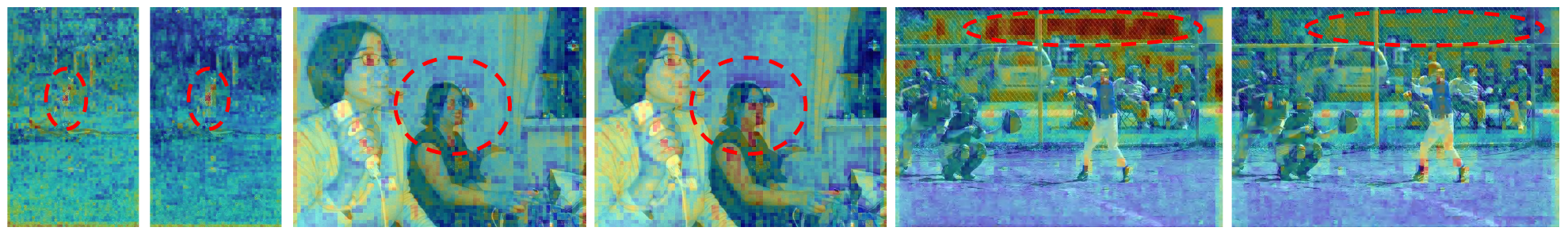}
        \put(1.5, -1.5){w/o \attentionname{}}
        \put(10.9, -1.5){w/ \attentionname{}}
        \put(25, -1.5){w/o \attentionname{}}
        \put(43, -1.5){w/ \attentionname{}}
        \put(63, -1.5){w/o \attentionname{}}
        \put(85, -1.5){w/ \attentionname{}}
    \end{overpic}
    \vspace{-10pt}
    \caption{
        Visualization of feature maps for method w/ and w/o \attentionname{}.
        We showcase examples from different images for each branch to further highlight the effectiveness of our \attentionname{} in improving the localization accuracy of feature extraction.
        The right part depicts the method with our \attentionname{}.
    }
    \vspace{-8pt}
    \label{fig:gql_vis}
\end{figure*}

\subsection{Rethinking Multi-Scale Feature Learning in Basic Building Blocks}
\label{sec:rethinking}

As mentioned in \secref{sec:intro}, redundancy exists among the features from different branches in the two popular blocks, namely CSP \cite{wang2020cspnet} and ELAN \cite{wang2022yolov7}, potentially compromising the multi-scale representation capability.
To compensate for this deficiency, we revisit the basic block design of previous real-time object detectors and adopt the hierarchical multi-branch structure \cite{gao2021res2net} instead of CSP and ELAN in our \fullblockname{}. 
As depicted in~\figref{fig:detailed_structure_compare}, compared with the two-branch structure in CSP and the parallel-like multi-branch structure in ELAN, the hierarchical multi-branch structure enables each branch of the basic block to have different receptive fields.
This hierarchical multi-branch structure is the key of Res2Net \cite{gao2021res2net} to augment the richness of information at different scales in building blocks.
%

The specifics of the multi-scale block of \fullblockname{} are illustrated in the right part of \figref{fig:detailed_structure_compare}(b).
It is utilized in the backbone and neck of the detector.
Let $\mathbf{Z} \in R^{H \times W \times C}$ be the input feature, where $C$, $H$, and $W$ are the channel number, width, and height of $\mathbf{Z}$, respectively.
We first use a $1 \times 1$ convolution to transform $\mathbf{Z}$
to $\mathbf{X} \in R^{H \times W \times NC}$.
Then, we split $\mathbf{X}$ into $N$ distinct groups, denoted as $\{\mathbf{X}_i\}$, where $i \in {1,2,3,...,N}$.
To achieve a better performance and speed trade-off, we opt for $N$ to be 3.
Except for $\mathbf{X}_1$, each group goes through a branch with inverted bottleneck module (IBM) \cite{sandler2018mobilenetv2} to attain the output $\mathbf{Y}_i$.
This is different from Res2Net in which a standard convolution is used in each branch.
Another advantage of IBM is the use of depthwise convolution, which makes using large-kernel convolution possible.
Note that a simplified inverted bottleneck module (SIBM) without the last 1$\times$1 layer is utilized in small-scale models for faster speed. 
The mathematical representation of $\mathbf{Y}_i$ can be described as follows:
\begin{equation}
  \mathbf{Y}_i = \left\{
  \begin{array}{lc}
    \mathbf{X}_i,  & i=1 \\
    F_I(\mathbf{Y}_{i-1} + \mathbf{X}_i), & i>1
  \end{array} \right. 
  \label{eq:eq_msblock}
\end{equation}
where $F_I$ denotes the (S)IBM module.
Finally, we concatenate all branches and apply a $1 \times 1$ convolution to interact among all branches, each of which encodes features at different scales.
This 1$\times$1 convolution is also used to adjust the channel numbers when the network goes deeper.

\subsection{Global Query Learning}
\label{sec:gql}

Though introducing the hierarchical structure mentioned above in building blocks can enable object detectors to capture rich multi-scale features, it neglects the importance of explicitly modeling the multi-scale features for objects at different scales.
The attention mechanism in Transformer has been proven to be an effective technique for establishing pairwise relationships between a given query $\mathbf{Q}$ and a key $\mathbf{K}$ \cite{vaswani2017attention, dosovitskiy2020image, guo2021attention_survey, han2022survey, khan2021transformers}.
Motivated by this, in this subsection, we propose an efficient strategy named \textbf{Global Query Learning} (\attentionname{}), aiming at offering cross-stage guidance for basic blocks.
As shown in \figref{fig:detailed_structure_compare}(b), we hold a lightweight and learnable global query, denoted as $\mathbf{Q} \in R^{N \times c^2}$, traveling different stages to convert global spatial information to each block to increase the diversity among branches.
$N$ and $c$ are the number of branches and the size of $\mathbf{Q}$, respectively.
%
It enables the spatial information $\mathbf{K}$ of the current block to index the weight, which is used for arranging the influence of each branch according to the stage where it is located and the information from the previous stage.
Note that we only use \attentionname{} in the $2^{nd}$, $3^{rd}$, and $4^{th}$ stages of the backbone for a better trade-off between computational cost and effectiveness.
Formally, as defined in \secref{sec:rethinking}, given $\mathbf{Y} \in R^{H \times W \times NC}$, the output $\mathbf{Y'} \in R^{H \times W \times NC} $ of our attention module can be formulated as
\begin{align}
  \mathbf{K} &= \mathrm{Linear}(\mathrm{GAP}(\mathbf{Y})), \\ 
  \mathbf{Y'} &= S(\mathbf{Q} \times \mathbf{K}^T) \odot  F_{ms}(\mathbf{Y}),
\end{align}
where $S$ represents the sigmoid function, GAP is the global average pooling operation, $\odot$ is element-wise multiplication, and $\mathrm{Linear}$ denotes a linear layer for extracting spatial information.
$F_{ms}$ denotes the multi-scale block designed to enhance inter-branch feature diversity, as depicted in \figref{fig:detailed_structure_compare}(b).
With $\mathrm{GAP}$, the global query $\mathbf{Q}$ is lightweight, enabling our proposed \methodname{} to have almost negligible computational cost.
In \figref{fig:gql_vis}, we present visualization evidence of some cases to demonstrate that \attentionname{} potentially reduces the negative impact of contaminative spatial information.
Further analyses of \attentionname{} are provided in \secref{sec:attention_analysis}.

\subsection{Heterogeneous Kernel Size Selection Protocol}

In addition to designing the building block, we also delve into the use of convolutions from a macro architecture perspective.
Although the building blocks we use bring large advancement on the multi-scale ability, they do not thoroughly explore the role of convolutions with different kernel sizes, especially for large-kernel ones, which has been shown effective in CNN-based models for visual recognition tasks \cite{hou2024conv2former,liu2022convnet,ding2022scaling} but are overlooked by real-time object detectors.  
The main obstacle in incorporating large-kernel convolutions into real-time object detectors is the computational overhead, especially for lower stages.
As shown in \tabref{tab:conv_time}, applying large-kernel convolutions to high-resolution features is computationally expensive, so adopting large-kernel convolutions for low-resolution features is better to reduce computational cost substantially.

Furthermore, previous real-time object detectors mostly adopt homogeneous convolutions (\ie, convolutions with the same kernel size) in different encoder stages, but we argue that this is not the optimal option for extracting multi-scale semantic information. 
In a pyramid architecture, high-resolution features extracted from the shallow stages of the detectors are commonly used to capture fine-grained semantics, which are used to detect small objects.
On the contrary, low-resolution features from deeper stages are utilized to capture high-level semantics, which are used to detect large objects.
If we adopt uniform small-kernel convolutions for all stages, the \textit{Effective Receptive Field (ERF)} in the deep stage is limited, affecting the performance on large objects.
Incorporating large-kernel convolutions in each stage can help address this limitation. 
Still, they are with a large ERF and encode broader regions, increasing the probability of including contaminative information outside the small objects.
 
According to the above analysis, we present a \textbf{Heterogeneous Kernel Size Selection} (\protocolname{}) Protocol.
\protocolname{} leverages heterogeneous convolutions in different stages to capture richer multi-scale features.
To be specific, we gradually increase the kernel size from lower stages to higher ones.
Since kernel sizes are typically odd numbers, we start from $3 \times 3$ and adopt a step size of 2 for increasing kernel size.
With this protocol, the kernel sizes of convolutions are 3, 5, 7, and 9 from the shallowest to the deepest stage.
\revised{
Different from previous works \cite{maaz2023edgenext, hou2024conv2former}, we also extend these settings into the PAFPN and Head parts.
In addition, thorough analyses are conducted to demonstrate why we introduced the \protocolname{} protocol and its effectiveness in~\secref{sec:hks_protocol}. 
}
Our \protocolname{} protocol can enlarge the receptive fields in the deep stages without introducing any other impact on the shallow stages.
It allows for extracting fine-grained and coarse-grained semantic information, enhancing the multi-scale feature representation capability.
It not only helps encode richer multi-scale features but also ensures efficient inference. 
In practice, we empirically found that our \methodname{} with \protocolname{} protocol achieves nearly the same inference speed as that only using $3\times3$ convolutions.

\begin{table}[ht]
        \centering
        \footnotesize
        \setlength{\tabcolsep}{6pt}
        \caption{
            FPS ($\times 10^3$) of convolutions with different kernel sizes in different stages of the network.
            The \textbf{gray} color indicates the kernel size used in each stage of our \methodname{}.
            The computation environment of the benchmark for all models is the same.
        }
        \vspace{-8pt}
        \begin{tabular}{ccccccc}
        Stage &  Input Size      & \#Channels& \conv{3}         & \conv{5}          & \conv{7}         & \conv{9}           \\ \midrule[0.8pt]
        \#1   &  $320\times320$  & 160       & \cc \textbf{2.72}& 1.38              & 0.93             & 0.65               \\ 
        \#2   &  $160\times160$  & 320       & 5.53             & \cc \textbf{2.78} & 1.86             & 1.31               \\ 
        \#3   &  $80\times80$    & 640       & 10.46            & 5.52              & \cc \textbf{3.65}& 2.65               \\ 
        \#4   &  $40\times40$    & 1280      & 14.25            & 10.73             & 7.21             &  \cc \textbf{5.21} \\
      \end{tabular}
      \label{tab:conv_time}
\end{table}

\subsection{Architecture}

The backbone of our model consists of four stages, each of which is followed by a $3 \times 3$ convolution with stride 2 for downsampling.
We adopt the \sArt real-time detector, RTMDet \cite{lyu2022rtmdet}, as our baseline. 
In our encoder, we utilize SiLU \cite{elfwing2018silu} as the activation function and BN \cite{ioffe2015bn} for normalization.
We add an SPP block \cite{he2015spp} after the third stage as done in \cite{lyu2022rtmdet}.
Following \cite{bochkovskiy2020yolov4, wang2020cspnet}, we use PAFPN as neck to construct a feature pyramid \cite{lin2017feature, liu2018pafpn} upon our encoder.
It fuses multi-scale features extracted from different stages in the backbone.
The fundamental building blocks employed in the neck are also our \fullblockname{}, and the \protocolname{} is similarly utilized in both the neck and head.
Additionally, to achieve a better trade-off between speed and accuracy, we halve the channel depth of the multi-level features from the backbone.
We present three variants of \methodname{}, namely \methodname{}-XS, \methodname{}-S, and \methodname{}. 
The detailed configurations of different scales of our \methodname{} are listed in \tabref{brief_config}.
For other aspects, we maintain consistency with \cite{lyu2022rtmdet}.

\begin{table*}[ht]
    \centering
    \footnotesize
    \setlength{\tabcolsep}{8pt}
        \caption{
            Brief configurations of the proposed \methodname{}.
            ``Module Type'' represents the type of module in the basic blocks.
            ``Widen Factor'' indicates a factor to scale the channel dimension.
            The base channel setting is $\{32, 64, 128, 512, 256\}$.
            ``Module Number'' is the number of basic blocks in the encoder. 
            Each of the three branches in \methodname{} contains two \fullblockname{}.
            ``Channel Expansion Ratio'' denotes the ratio of the channel expansion process within the IBM of the \fullblockname{}.
            $c$ refers to the channel dimension of the IBM's input.
            Based on RTMDet, we implement three variants with parameters 4.54M, 8.13M, and 22.17M, respectively.
        }
        \vspace{-8pt}
        \begin{tabular}{lccccccc}
            Model             & Module Type     & Widen Factor  & Module Number     & Channel Expansion Ratio  & Latency    & Params   & MACs    \\ \midrule[0.8pt]
            RTMDet-XS         & CSPNext Module  & 0.375         & 1 + 1 + 1 + 1     & -                        & 6.5ms      & 4.9M     & 8.1G     \\ 
            \methodname{}-XS  & SIBM + SIBM   & 1.050         & 2 + 2 + 2 + 2     & $c \rightarrow 2c$       & 7.1ms      & 5.1M     & 8.7G     \\ \midrule[0.8pt]
            RTMDet-S          & CSPNext Module  & 0.500         & 1 + 2 + 2 + 1     & -                        & 7.3ms      & 8.9M     & 14.8G    \\ 
            \methodname{}-S   & SIBM + SIBM   & 1.375         & 2 + 2 + 2 + 2     & $c \rightarrow 2c$       & 7.3ms      & 8.7M     & 15.0G    \\ \midrule[0.8pt]
            RTMDet-M          & CSPNext Module  & 0.750         & 2 + 4 + 4 + 2     & -                        & 10.0ms     & 24.7M    & 39.3G    \\ 
            \methodname{}     & SIBM + IBM   & 2.175         & 2 + 2 + 2 + 2     & $c \rightarrow 2c$       & 10.5ms     & 23.3M    & 38.8G    \\ 
        \end{tabular}%
        \label{brief_config}
\end{table*}

\section{Experiments} \label{sec:exp}

\subsection{Experiment Setup}

\myPara{Implementation Details.}
Our implementation is based on the MMDetection \cite{mmdetection} framework and PyTorch \cite{paszke2019pytorch}.
All experiments are conducted with a batch size of 32 per GPU to guarantee impartiality on a machine with 8 NVIDIA 3090 GPUs. 
To be specific, due to the hardware limitation, the batch size for the large version model is 16 per GPU.
All the scales of \methodname{} are trained from scratch for 300 epochs without relying on other large-scale datasets, like ImageNet \cite{deng2009imagenet}, or pre-trained weights.
The input size of all the experiments is $640 \times 640$.
During training, we employ the AdamW optimizer \cite{kingma2017adam} with a momentum of 0.9 and a weight decay of 0.05. 
The weight decay is set to 0 for bias and normalization parameters.
The learning rate setting includes a 1000-iteration warm-up with a start factor of $1 \times 10^{-5}$ and a flat cosine annealing schedule with an initial value of $1 \times 10^{-4}$. 
\revised{
We also incorporate the Exponential Moving Average (EMA) \cite{klinker2011exponential} with a decay factor of 0.9998 to enhance the model's performance.
}
All the experiments are trained from scratch for 300 epochs.
Furthermore, we utilize Focal Loss \cite{lin2017focal} and DIoU Loss \cite{rezatofighi2019generalized} for classification and bounding box regression.
The label assignment is based on SimOTA \cite{ge2021yolox}, which incorporates a cost function for matching that includes classification cost, region prior cost, and regression cost.
Further details can be found in the work of Lyu et al. \cite{lyu2022rtmdet}.
For data augmentation, we use the cached Mosaic \cite{lyu2022rtmdet} and MixUP \cite{zhang2018mixup} in the first 280 epochs and LSJ \cite{ghiasi2021ljs} in the last 20 epochs.
For a fair comparison, during evaluation, we ensure the consistency of post-processing with the previous practice \cite{jocher2020yolov5,li2022yolov6,wang2022yolov7, lyu2022rtmdet}.
During post-processing, we filter out the bounding boxes with scores lower than 0.001 before non-maximum suppression (NMS), and the top 300 boxes are selected for evaluation.

\myPara{Datasets.}
We evaluate the proposed detector on the widely used MS COCO \cite{lin2014microsoft} benchmark, following the standard practice in most previous works. 
Specifically, we use the \textit{train2017} set, which includes 115\textit{K} images, for training, and the \textit{val2017} set, which includes 5\textit{K} images, for validation. 
The standard COCO-style measurement, i.e., Average Precision (\AP{}), is utilized as the primary challenge metric for evaluation.
In addition, the mAP with IoU thresholds of 0.5 and 0.75 and AP for small, medium, and large objects are reported as supporting metrics.

\myPara{Benchmark Settings.}
Following previous works, we measure all models' frames per second (FPS) using an NVIDIA 3090 GPU in a full-precision floating-point format (FP32). 
During testing, we perform inference without the NMS post-processing step, 
The batch size used for the inference process is set to 1. 
Additionally, MACs are calculated based on an input size of $640 \times 640$, using MMDetection Framework \cite{mmdetection}.

\begin{figure}[t]
    \centering
    \footnotesize
    \begin{overpic}
        [width=0.48\textwidth]{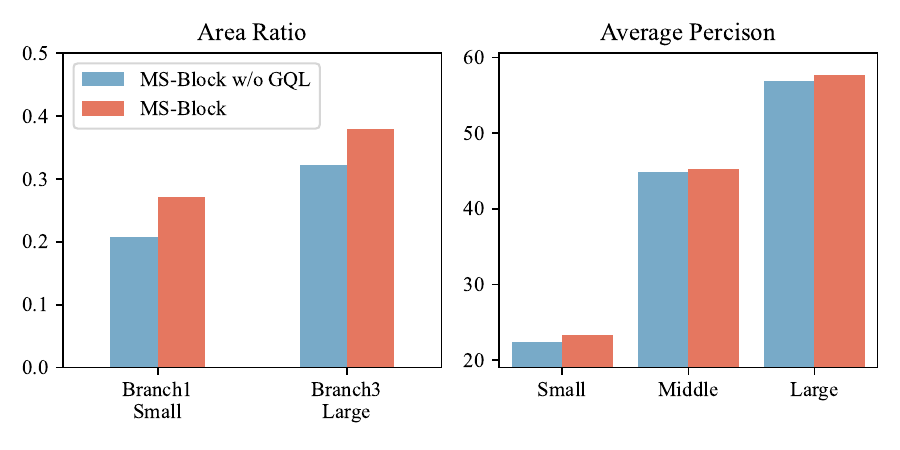}
        \put(27,0){(a)}
        \put(76,0){(b)}
    \end{overpic}
    \vspace{-8pt}
    \caption{
        (a) Area ratio of high feature activation value (>0.5) within the GT box for the branch corresponding to small and large objects.
        (b) Average Precision comparison for objects at different scales.
        The \textbf{red} color represents the method with \attentionname{}.
    }
    \label{fig:spatial_error}
\end{figure}

\begin{figure}[t]
    \centering
    \footnotesize
    \begin{overpic}
        [width=0.48\textwidth]{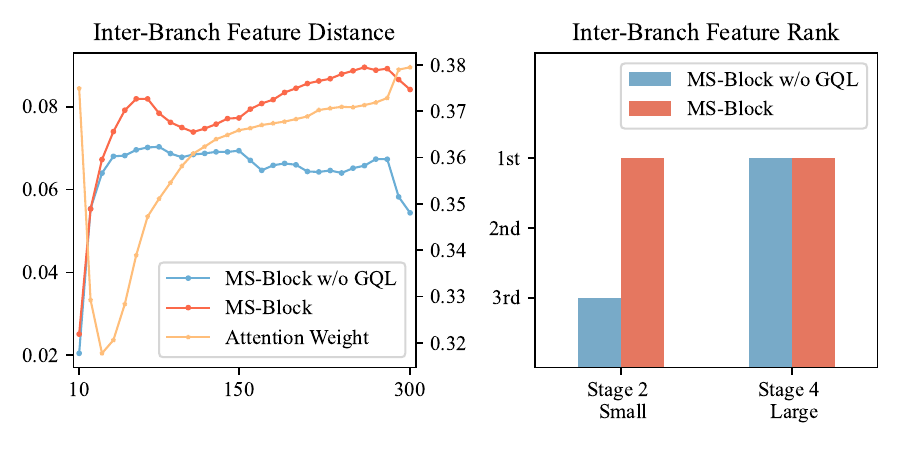}
        \put(25,0){(a)}
        \put(76,0){(b)}
    \end{overpic}
    \vspace{-8pt}
    \caption{
        (a) Trends of the average feature distances between the target branch and other branches among different stages. 
        (b) Ranking of the feature distribution of the target branch for the stage corresponding to the small and large objects. 
        The \textbf{red} color represents the method with \attentionname{}.
        The \textbf{yellow} line represents the trend of the target branch's attention weight.
    }
    \label{fig:attention_rank}
\end{figure}

\subsection{Analysis of \attentionname{}}
\label{sec:attention_analysis}
To validate the effectiveness of our \attentionname{}, 
we conduct a series of analyses in this subsection.
In \figref{fig:spatial_error}(a), 
we compute the area ratio of high activation features (value > 0.5) within GT boxes of small and large objects for the first branch with the smallest receptive field in the 2$^{nd}$ stage and the last branch with the largest receptive field in 4$^{th}$ stage, respectively.
\figref{fig:spatial_error}(a) shows that the feature map has more influential activation values within the regions of objects at the scale corresponding to the stage and branch where the feature map is extracted.
This implies that \attentionname{} can enable models to more accurately locate objects at different scales from a feature learning perspective.
In \figref{fig:spatial_error}(b), we present the average precision of different object scales. 
It intuitively shows that our \attentionname{} brings an obvious improvement in multi-scale performance.

Besides, in \figref{fig:attention_rank}(a), we visualize the trend of inter-branch feature distance and the attention weight of \attentionname{} during training.
For simplicity, we demonstrate the change using the average $L_1$ distance between the target branch and other branches.
As defined in \secref{sec:gql}, each value in \attentionname{} represents the weight of each branch within the \fullblockname{}.
For convenience, we only visualize the changing trend of the target branch's corresponding weight value, representing the target branch's influence among the block. 
The target branch we choose is based on the branch's receptive field and the stage, \eg the left branch is the target in 2$^{nd}$ stage, corresponding to small objects.
We take the average result of all the stages for comparison.
It is also evident that the attention weight of the target branch follows the same trend as the feature distance, further demonstrating the efficacy of \attentionname{}.
In \figref{fig:attention_rank}(b), we compare the average activation value of the feature map between each branch in different stages and get the ranking value.
With \attentionname{}, the left branch with a small receptive field exhibits a large feature activation in the shallow stage, and vice versa.
This indicates that \attentionname{} adaptively rearranges the effect of each branch in the corresponding stage to maximize its influence.

\subsection{Analysis of \protocolname{} Protocol}
\label{sec:hks_protocol}
Previous studies \cite{luo2016erf, ding2022scaling} have introduced the concept of ERF as a metric to understand the behavior of deep convolutional neural networks (CNNs). 
ERF measures the effective region in the input space influenced by a feature representation.
In this subsection, we leverage the concept of ERF further to investigate the effectiveness of \protocolname{}.
We follow the methodology outlined in RepLKNet \cite{ding2022scaling} to measure the ERF through the aggregated contribution score matrix, denoted as $\mathbf{A} \in R^{H \times W}$.
Suppose the input is $\mathbf{I} \in R^{H \times W \times C}$ and the output feature map is $\mathbf{F} \in R^{ H' \times W' \times C'}$, the computation process of $\mathbf{A}$ can be mathematically described as \cite{ding2022scaling}:
\begin{align}
    \mathbf{P} &= \max{(\sum_c^{C'}{\frac{\partial \mathbf{F} (\frac{H'}{2},\frac{W'}{2},c)}{\partial \mathbf{I}}}, 0)}, \\
    \mathbf{A} &= \log_{10} (\sum_c^C \mathbf{P}(:,:,c)+1).
\end{align}
Initially, we calculate $\mathbf{A}$ for stage 2, stage 3, and stage 4 in the encoder. 
Subsequently, each stage's $\mathbf{A}$ is normalized to [0, 1].
Suppose there is a threshold represented by $\theta$. 
The area ratio whose value is higher than $\theta$, \ie the high-contribution area, can be described as $h(\theta)$:
\begin{gather}
  h(\theta) = \frac{1}{|\mathbf{A}|} \sum_{a \in \mathbf{A}} \mathbbm{1}[a>\theta] ,
\end{gather}
where $\mathbbm{1}$ refers to the indicate function.
To intuitively show the difference between each model, we calculate $h$ under different $\theta$ from 0.5 to 0.9 with a step of 0.05 and take the average $\bar{h}$ to measure the ERF.

\begin{figure}[ht]
    \centering
    \begin{minipage}[t]{0.24\textwidth}
    \centering
    \begin{overpic}[width=\textwidth]{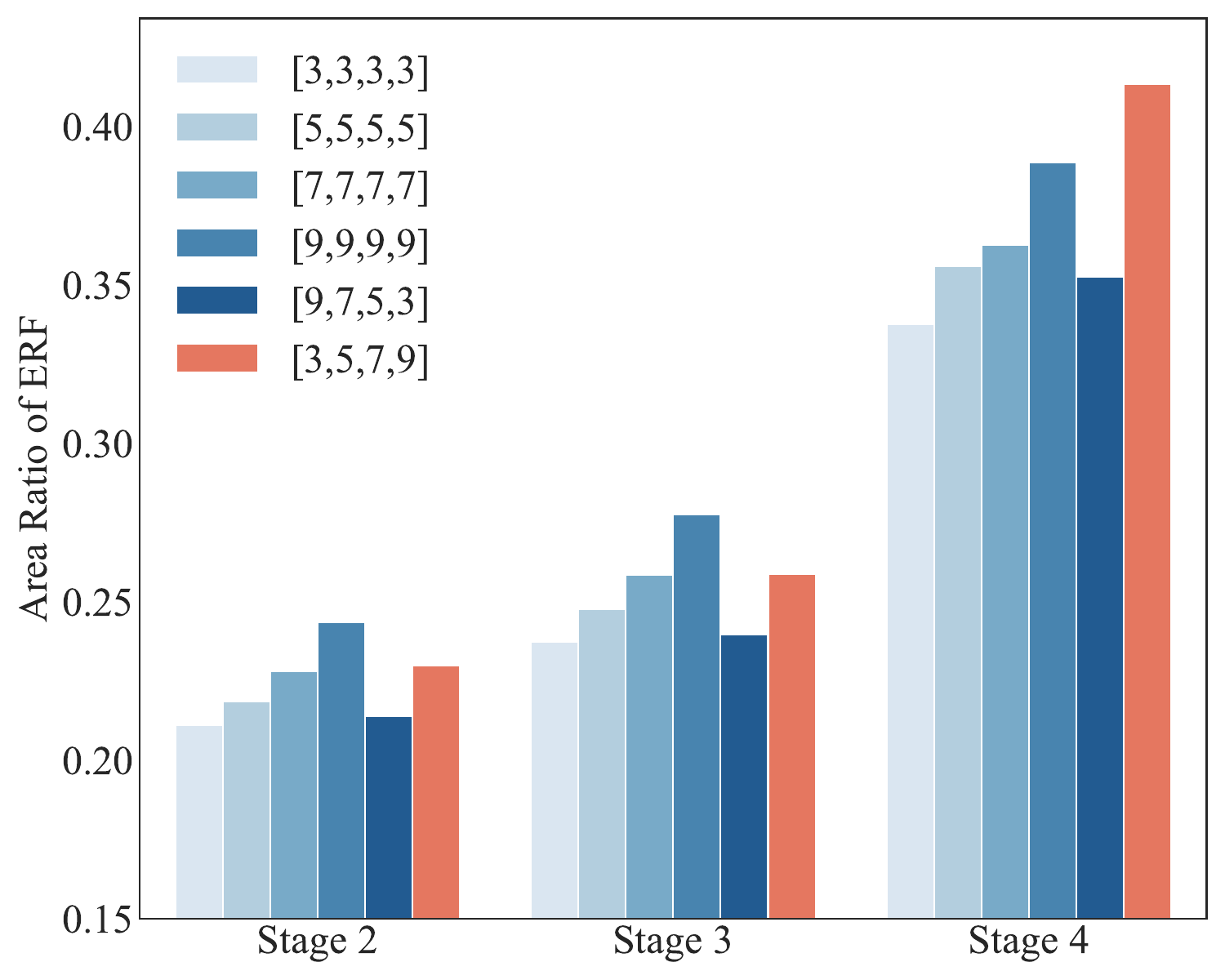}
        \put(52,-4){~\small{(a)}}
    \end{overpic}
    \end{minipage}
    \begin{minipage}[t]{0.24\textwidth}
    \centering
    \begin{overpic}[width=\textwidth]{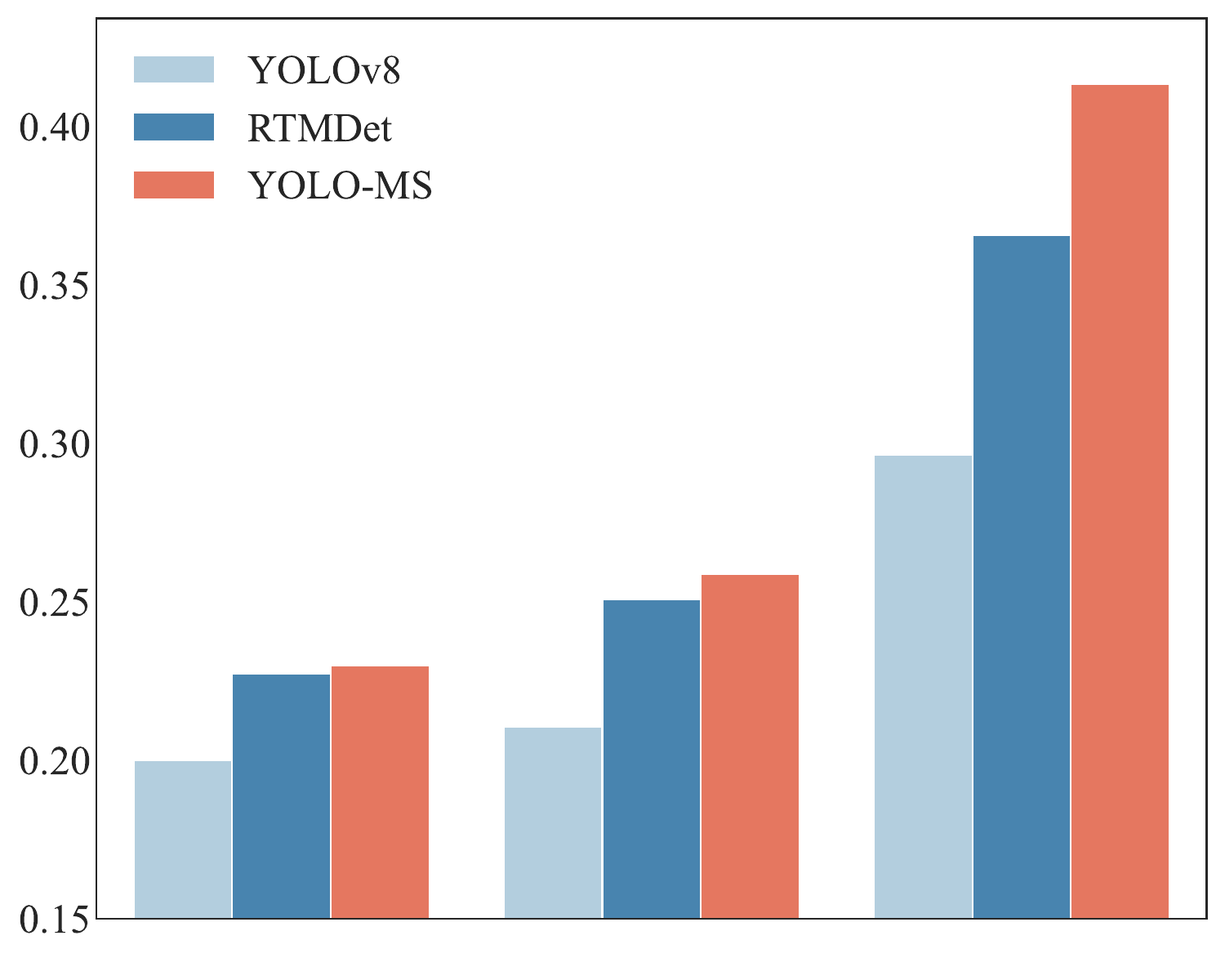}
        \put(52,-4){~\small{(b)}}
    \end{overpic}
    \end{minipage}
    \vspace{-10pt}
    \caption{
        Statistical analysis of the effective receptive field.
        (a) Comparison between different kernel size settings.
        (b) Comparison between different real-time detectors.
        $k_i$ represents the kernel size in the $i$th stage.
        The \textbf{red} color represents the method with \protocolname{}.
    }
    \label{fig:erf}
\end{figure}

The visual comparison is presented in \figref{fig:erf}.
To simplify notation, we use the format $[k_1,k_2,k_3,k_4]$, where $k_i$ represents the kernel size in stage $i$. 
As shown in \figref{fig:erf}(a), as the kernel size increases, the area of ERF also becomes larger in all stages, which supports the positive correlation between kernel size and receptive fields.
Moreover, the area of ERF is smaller than most other settings in the shallow stage, while in the deep stage, it is the opposite.
This observation indicates that the protocol effectively enlarges the receptive fields in the deep stages without compromising the shallow ones.
In \figref{fig:erf}(b), we can observe that our \protocolname{} achieves the best ERF in deep stages, allowing us to detect large objects better.

\subsection{Ablation Study}

\myPara{Ablations on the proposed methods.}
To investigate the impact of our proposed methods, 
we conduct an ablation study on gradually increasing components from the baseline RTMDet \cite{lyu2022rtmdet} to \methodname{}.
The results are reported in \tabref{tab:entire_ablation_study}. 
Intriguingly, the proposed \blockname{} with \attentionname{} and \protocolname{} leads to all-around improvement to the model.
Specifically, our proposed methods bring a significant improvement of +1.8\% AP compared to RTMDet. 

\begin{table}[ht]
    \centering
    \footnotesize
    \setlength{\tabcolsep}{2.5pt}
        \caption{
            Ablation study on the proposed methods.
            All models are tiny versions and trained from scratch.
            The baseline is RTMDet-T without pre-trained weight.
            All models are benchmarked under the same computational environment.
        }
        \vspace{-10pt}
        \begin{tabular}{cccccccccc}
            \fullblockname{} & +\attentionname{} &  +\protocolname{} & \AP{} &\AP{s} & \AP{m}&\AP{l}  & FPS & Params & MACs   \\ \midrule[0.8pt]
            \multicolumn{3}{c}{RTMDet-T}                             & 40.3  &  20.9 &  44.8 &  57.4  & 154 & 4.9M   & 8.1G   \\ \midrule[0.8pt]
            \multicolumn{3}{c}{Res2Net}                              & 40.0  &  21.3 &  44.8 &  55.2  & 170 & 4.5M   & 8.2G   \\  
            \cmark{}         &                   &                   & 41.0  &  22.4 &  45.2 &  56.9  & 159 & 4.2M   & 8.6G   \\ 
                             &                   & \cmark{}          & 41.3  &  22.7 &  45.3 &  57.4  & 158 & 4.2M   & 8.6G   \\ 
            \cmark{}         &  \cmark{}         &                   & 41.5  &  23.3 &  45.7 &  57.7  & 150 & 4.2M   & 8.6G   \\ \rowcolor[gray]{.95}
            \cmark{}         &  \cmark{}         & \cmark{}          & 42.8  &  23.1 &  46.8 &  60.1  & 141 & 5.1M   & 8.7G   \\ 
        \end{tabular}
        \label{tab:entire_ablation_study}
\end{table}

\begin{figure*}[ht]
    \centering
    \begin{overpic}
        [width=0.8\textwidth]{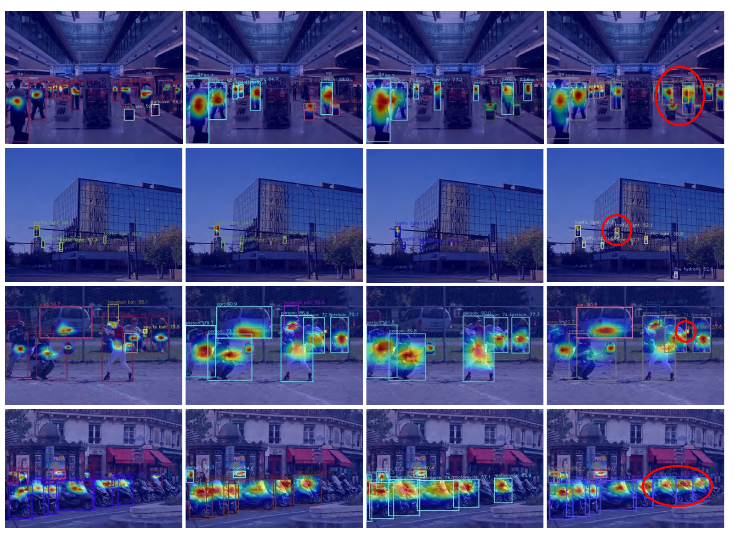}
        \put(7, -1){ ~\small{YOLOv7~\cite{wang2022yolov7}}}
        \put(31, -1){ ~\small{RTMDet~\cite{lyu2022rtmdet}}}
        \put(57, -1){ ~\small{YOLOv8~\cite{jocher2023yolov8}}}
        \put(81, -1){ ~\small{YOLO-MS}}
    \end{overpic}
    \vspace{-5pt}
    \caption{
        Visual comparison with \sArt{} through Grad-CAM \cite{selvaraju2017gradcam}.
        Our methods can better locate the objects at different scales.
    }
    \label{fig:cam}
\end{figure*}
 
\myPara{Ablations on branch number.}
Our \blockname{} partitions and propagates the input feature through multiple branches. 
However, increasing the branch number also increases IBM and decreases the channel number in each branch. 
To investigate the impact of the number of branches, 
denoted as $N_b$, we conduct an ablation study here.
The results are detailed in \tabref{tab:number_of_branches}. 
To achieve a better performance and speed trade-off, we designate $N_b=3$ as the default setting in all subsequent experiments.
If the speed is not an obstacle, $N_b=4$ is also recommended.

\begin{table}[ht]
        \footnotesize
        \centering
        \setlength{\tabcolsep}{6pt}
        \caption{
            Ablation study on the branch number of \blockname{}.
            $N_b$ refers to the number of branches in the \blockname{}.
            The baseline is \methodname{}-XS without \protocolname{} protocol.
            All models are benchmarked under the same computational environment.
        }
        \vspace{-10pt}
        \begin{tabular}{lcccccccc}
                $N_b$    & \AP{} &\AP{s} & \AP{m} &\AP{l} & FPS  & Params& MACs     \\ \midrule[0.8pt]
                2        & 39.0  &  20.9 &  43.5  & 54.2  & 184  & 3.5M  & 7.6G     \\ \rowcolor[gray]{.95}
                3        & 41.5  &  23.3 &  45.7  & 57.7  & 150  & 4.2M  & 8.6G     \\ 
                4        & 43.5  &  24.0 &  47.3  & 60.3  & 137  & 5.1M  & 9.7G     \\ 
        \end{tabular}
        \label{tab:number_of_branches}
\end{table}

\myPara{Ablations on the spatial dimension of the query.}
Our \attentionname{} maintains a global query to provide cross-stage information for adaptive augmenting the branch features.
We conduct an ablation study to investigate the impact of the spatial dimension of the query.
Intriguingly, directly increasing the size of branches does not always lead to performance improvement.  
The results are summarized in \tabref{tab:ablation_study_query_size}. 
Specifically, when $D_s=4$, \methodname{} reaches the best performance of 41.5\% AP.
Considering the performance and speed trade-off, we choose $4^2$ as the spatial dimension of the global query in all subsequent experiments.

\begin{table}[ht]
    \centering
    \scriptsize
    \setlength{\tabcolsep}{8pt}
        \caption{
            Ablation study on the spatial dimension of Global Query.
            $D_s$ refers to the spatial dimension of the global sharing query in \attentionname{}.
            The baseline is \methodname{}-XS without \protocolname{} protocol and trained from scratch.
            All models are benchmarked under the same computational environment.
        }
        \vspace{-10pt}
        \begin{tabular}{lccccccc}
            $D_s$       & \AP{}  &\AP{s} & \AP{m}  & \AP{l}   & FPS   & Params & MACs \\ \midrule[0.8pt]
            $2^2$       & 41.0   & 22.2  & 45.3    &  57.0    & 153   & 4.2M   & 8.6G \\ \rowcolor[gray]{.95}
            $3^2$       & 41.2   & 22.4  & 45.5    &  57.4    & 152   & 4.2M   & 8.6G \\ 
            $4^2$       & 41.5   & 23.3  & 45.7    &  57.7    & 150   & 4.2M   & 8.6G \\ 
            $5^2$       & 41.4   & 22.2  & 45.6    &  57.6    & 145   & 4.2M   & 8.6G \\ 
            $6^2$       & 41.4   & 22.7  & 46.0    &  57.4    & 142   & 4.2M   & 8.6G \\ 
        \end{tabular}
        \label{tab:ablation_study_query_size}
\end{table}

\begin{table}[ht]
        \centering
        \footnotesize
        \setlength{\tabcolsep}{3pt}
        \caption{
            Comparison with different kernel size settings of \protocolname{}.
            $k_i$ represents the kernel size in the $i$th stage.
            `Neck-HKS' and `Head-HKS' refer to using HKS protocol in PAFPN and Head module. 
            The baseline is YOLO-MS-XS.
            All models are trained from scratch and benchmarked under the same computational environment.
        }
        \vspace{-10pt}
        \begin{tabular}{lcccccccccc}
         $[k_1, k_2, k_3, k_4]$  & \AP{}  & \AP{s}   & \AP{m}  & \AP{l}  & Params     & MACs    \\ \midrule[0.8pt]
         $[3,3,3,3]$             & 41.0   & 22.4     & 45.2    & 56.9    & 4.2M       & 8.6G    \\ 
         $[5,5,5,5]$             & 41.7   & 22.7     & 46.2    & 57.7    & 4.3M       & 8.7G    \\ 
         $[7,7,7,7]$             & 41.8   & 23.4     & 46.2    & 58.4    & 4.3M       & 8.9G    \\
         $[9,9,9,9]$             & 41.8   & 22.3     & 46.4    & 57.8    & 4.4M       & 9.1G    \\ 
         $[11,11,11,11]$         & 41.9   & 22.7     & 46.8    & 57.7    & 4.5M       & 9.6G    \\ \midrule[0.8pt]
         $[5,7,9,11]$            & 41.9   & 22.7     & 46.8    & 57.7    & 4.5M       & 9.6G    \\ 
         $[3,7,11,15]$           & 41.9   & 22.7     & 46.8    & 57.7    & 4.5M       & 9.6G    \\ 
         $[9,7,5,3]$             & 41.2   &  22.0    & 45.2    & 58.2    & 4.3M       & 9.0G    \\ 
         $[3,5,7,9]$             & 41.9   &  22.6    &  46.2   & 58.9    & 4.2M       & 8.6G    \\ \midrule[0.8pt]
         + Neck-HKS              & 42.3   &  22.0    & 46.1    & 59.8    & 4.4M       & 8.7G    \\ 
         + Neck-HKS + Head-HKS   & 42.8   &  23.1    &  46.8   & 60.1    & 5.1M       & 8.7G    \\ 
        \end{tabular}%
        \label{tab:hks}
\end{table}

\begin{table}[t]
    \centering
    \footnotesize
    \setlength{\tabcolsep}{8pt}
    \caption{
        Comparison with different kernel size settings.
        ``Ours'' represents \methodname{}-XS.
        ``TTA'' refers to the Test Time Augmentation.
    }
    \vspace{-10pt}
    \begin{tabular}{lcccccc}
     Model             &  Resolution                       & \AP{}   & \AP{s}  & \AP{m}   & \AP{l}  \\ \midrule[0.8pt]
     RTMDet-tiny       &  \multirow{2}*{$320\times320$}    & 30.0    & 8.3     & 36.0     & 54.0    \\ 
     Ours              &                                   & 32.7    & 10.5    & 35.3     & 56.8     \\ \midrule[0.8pt]
     RTMDet-tiny       &  \multirow{2}*{$640\times640$}    & 41.0    & 20.7    & 45.3     & 58.0    \\ 
     Ours              &                                   & 42.8    & 23.1    & 46.8     & 60.1    \\ \midrule[0.8pt]
     RTMDet-tiny       &  \multirow{2}*{$1280\times1280$}  & 35.2    & 29.2    & 45.8     & 36.4    \\ 
     Ours              &                                   & 37.9    & 31.0    & 46.7     & 37.5    \\ \midrule[0.8pt]
     RTMDet-tiny       &  \multirow{2}*{TTA}               & 41.9    & 28.1    & 47.1     & 55.5    \\ 
     Ours              &                                   & 45.2    & 31.0    & 48.4     & 60.6    \\
    \end{tabular}%
    \label{tab:different_resolution}
    \vspace{-10pt}
\end{table}

\begin{table*}[t]
    \centering
    \footnotesize
    \setlength{\tabcolsep}{12pt}
    \caption{
        Comparison with the of \sArt real-time object detectors.
        ``$^\dagger$'' refers to train with pertained models.
        All models are benchmarked under the same computational environment.
        \revised{
        The input size for evaluating ATSS, GFocalv2, and TOOD is $1333 \times 800$. 
        For evaluating other detectors, it is $640 \times 640$.
        The MACs computation for all models is under the input size of $640\times640$.
        The latency computation is without NMS.
        The inference of YOLOv10 is with one-to-one training.
        }
        The performance results of other detectors are referenced from MMDetection \cite{mmdetection} and the office repository.
        \revised{Note that EMA, which is a common technique for model improvement, is used for all the training of YOLO models and RT-DETR. }
    }
    \vspace{-10pt}
    \renewcommand\arraystretch{1.0}
    \begin{tabular}{lcccccccccc}
        Model                                        & \AP{} &\AP{50} &\AP{75}&\AP{s}  & \AP{m} &\AP{l} & Latency   & Params  & MACs    \\ \midrule[0.8pt]
        \revised{ATSS-R50 $^\dagger$ \cite{zhang2020bridging}}  & 39.4  & 57.6   & 42.8  & 23.6   & 42.9   & 50.3  &  -        & 32.3M   & 82.0G   \\
        \revised{GFocalv2-R50 $^\dagger$ \cite{li2021gfocalv2}} & 40.2  & 58.4   & 43.3  & 23.3   & 44.0   & 52.2  &  -        & 32.4M   & 83.3G   \\
        \revised{TOOD-R50 $^\dagger$ \cite{feng2021tood}}       & 42.4  & 59.7   & 46.2  & 25.4   & 45.5   & 55.7  &  -        & 32.2M   & 80.4G   \\ \midrule[0.8pt]
        YOLOv5-N \cite{jocher2020yolov5}                  & 28.0  & 45.9   & 29.4  & 14.0   & 31.8   & 36.6  & 4.8ms     & 1.9M    & 2.3G    \\ 
        YOLOv5-S \cite{jocher2020yolov5}                  & 37.7  & 57.1   & 41.0  & 21.7   & 42.5   & 48.8  & 5.2ms     & 7.2M    & 8.3G    \\ 
        YOLOv5-M \cite{jocher2020yolov5}                  & 45.3  & 64.1   & 49.4  & 28.4   & 50.8   & 57.7  & 7.1ms     & 21.2M   & 24.5G   \\  
        YOLOv5-L \cite{jocher2020yolov5}                  & 48.8  & 67.4   & 53.3  & 33.5   & 54.0   & 61.8  & 9.4ms     & 46.6M   & 54.6G   \\ \midrule[0.8pt]
        YOLOv6-N \cite{li2022yolov6}                 & 36.2  & 51.6   & 39.2  & 16.8   & 40.2   & 52.6  & 7.9ms     & 4.3M    & 5.5G    \\ 
        YOLOv6-T \cite{li2022yolov6}                 & 40.3  & 57.4   & 43.9  & 21.2   & 45.6   & 57.5  & 8.0ms     & 9.7M    & 12.4G   \\ 
        YOLOv6-S \cite{li2022yolov6}                 & 43.7  & 60.8   & 47.0  & 23.6   & 48.7   & 59.8  & 8.5ms     & 17.2M   & 21.9G   \\ 
        YOLOv6-M \cite{li2022yolov6}                 & 48.4  & 65.7   & 52.7  & 30.0   & 54.1   & 64.5  & 14.2ms    & 34.3M   & 40.7G   \\ 
        YOLOv6-L \cite{li2022yolov6}                 & 51.0  & 68.4   & 55.2  & 33.5   & 56.2   & 67.3  & 14.3ms    & 58.5M   & 71.4G   \\ \midrule[0.8pt]
        YOLOv7-T \cite{wang2022yolov7}               & 37.5  & 55.8   & 40.2  & 19.9   & 41.1   & 50.8  & 4.9ms     & 6.2M    & 6.9G    \\ 
        YOLOv7-L \cite{wang2022yolov7}               & 50.9  & 69.3   & 55.3  & 34.7   & 55.1   & 66.6  & 10.4ms    & 36.9M   & 52.4G   \\ \midrule[0.8pt]
        RTMDet-T$^\dagger$ \cite{lyu2022rtmdet}      & 41.0  & 57.4   & 44.4  & 20.7   & 45.3   & 58.0  & 6.5ms     & 4.9M    & 8.1G    \\ 
        RTMDet-S$^\dagger$ \cite{lyu2022rtmdet}      & 44.6  & 61.7   & 48.3  & 24.2   & 49.2   & 61.8  & 7.3ms     & 8.9M    & 14.8G   \\ 
        RTMDet-M \cite{lyu2022rtmdet}                & 49.3  & 66.9   & 53.9  & 30.5   & 53.6   & 66.1  & 10.0ms    & 24.7M   & 39.3G   \\ \midrule[0.8pt] 
        Gold-YOLO-N \cite{wang2023goldyolo}          & 39.6  & 55.7   &  –    & 19.7   & 44.1   & 57.0  &  -        & 5.6M    & 6.0G    \\  
        Gold-YOLO-S \cite{wang2023goldyolo}          & 45.4  & 62.5   &  –    & 25.3   & 50.2   & 62.6  &  -        & 21.5M   & 23.0G   \\  
        Gold-YOLO-M \cite{wang2023goldyolo}          & 49.8  & 67.0   &  –    & 32.3   & 55.3   & 66.3  &  -        & 41.3M   & 43.0G   \\  \midrule[0.8pt] 
        YOLOv8-N \cite{jocher2023yolov8}             & 37.2  & 52.7   & 40.3  & 18.9   & 40.5   & 52.5  & 5.3ms     & 3.2M    & 4.4G    \\
        YOLOv8-S \cite{jocher2023yolov8}             & 43.9  & 60.8   & 47.6  & 25.3   & 48.7   & 59.5  & 5.8ms     & 11.2M   & 14.4G   \\
        YOLOv8-M \cite{jocher2023yolov8}             & 49.8  & 66.9   & 54.2  & 32.6   & 54.9   & 65.9  & 8.3ms     & 25.9M   & 39.6G   \\  \midrule[0.8pt] 
        \revised{YOLOv9-T \cite{wang2024yolov9}}     & 37.5  & 52.3   & 40.6  & 18.4   & 41.7   & 52.9  & 10.3ms    & 2.1M    & 4.1G    \\
        \revised{YOLOv9-S \cite{wang2024yolov9}}     & 46.8  & 63.4   & 50.7  & 26.6   & 56.0   & 64.5  & 10.5ms    & 7.1M    & 13.4G   \\ \midrule[0.8pt] 
        \revised{YOLOv10-N \cite{wang2024yolov10}}   & 38.5  & 53.8   & 41.7  & 19.0   & 42.3   & 54.7  & 6.3ms     & 2.8M    & 4.3G    \\
        \revised{YOLOv10-S \cite{wang2024yolov10}}   & 46.2  & 62.9   & 50.2  & 26.8   & 51.0   & 63.6  & 6.9ms     & 8.1M    & 12.4G   \\
        \revised{YOLOv10-M \cite{wang2024yolov10}}   & 51.0  & 68.0   & 55.7  & 33.7   & 56.3   & 66.9  & 8.8ms     & 16.5M   & 31.5G   \\  \midrule[0.8pt] 
        \revised{RT-DETR-R18 $^\dagger$ \cite{lv2023rtdetr}}    & 46.5  & 63.8   & 50.4  & 28.4   & 49.8   & 63.0  & 9.5ms     & 20.0M   & 30.0G   \\  
        \revised{RT-DETR-R34 $^\dagger$ \cite{lv2023rtdetr}}    & 48.9  & 66.8   & 52.8  & 30.9   & 52.3   & 66.3  & 12.3ms    & 31.0M   & 45.1G   \\  \midrule[0.8pt] 
        \methodname{}-XS                             & 42.8  & 60.0   & 46.7  & 23.1   & 46.8   & 60.1  & 7.1ms     & 5.1M    & 8.7G    \\ 
        \methodname{}-S                              & 45.4  & 62.8   & 49.5  & 25.9   & 49.6   & 62.4  & 7.3ms     & 8.7M    & 15.0G   \\ 
        \methodname{}                                & 49.7  & 67.2   & 54.0  & 32.8   & 53.8   & 65.6  & 10.5ms    & 23.3M   & 38.8G   \\  \midrule[0.8pt] 
        YOLOv8-MS-N                                  & 40.2  & 56.5   & 43.3  & 20.9   & 44.1   & 55.5  & 6.5ms     & 2.9M    & 4.4G    \\ 
        YOLOv8-MS-S                                  & 46.2  & 63.3   & 50.1  & 27.0   & 51.0   & 62.7  & 6.9ms     & 9.5M    & 13.3G   \\  
        YOLOv8-MS-M                                  & 50.6  & 67.8   & 55.1  & 33.6   & 55.8   & 65.7  & 9.3ms     & 25.9M   & 35.2G   \\ \midrule[0.8pt] 
        \revised{YOLOv9-MS-T}                        &  38.5 & 53.7   & 41.9  & 19.3   & 42.8   & 52.8  & 6.1ms     & 2.0M    & 4.2G    \\  
        \revised{YOLOv10-MS-S}                       &  46.8 & 63.6   & 51.1  & 27.7   & 51.3   & 62.5  & 7.7ms     & 7.1M    & 11.5G   \\  
    \end{tabular}
    \label{tab:comparison_sota}
\end{table*}
\vspace{-5pt}

\myPara{Ablations on different kernel settings.}
We perform a quantitative comparison using different kernel size settings
to evaluate the effectiveness of \protocolname{}.
We explore both homogeneous kernel size settings with 3, 5, 7, 9, and 11 and an inverted version of \protocolname{}, \ie $[9,7,5,3]$.
As shown in \tabref{tab:hks}, simply increasing the kernel size improves the performance but brings more computational cost, impeding the efficiency of inference.
Furthermore, the order in which the kernels are arranged within the stages plays a crucial role.
Specifically, when a large-kernel convolution is used in shallow stages and a small-kernel one in deep stages, the performance drops by 0.7\% AP compared to \protocolname{}. 
This result suggests that, unlike the shallow stage, the deep stage requires a larger receptive field to capture coarse-grained information effectively.
Considering the computational cost, our \protocolname{} stands out as it incurs the least computation overhead.
This indicates that by strategically placing convolutions with different kernel sizes in suitable positions, we can maximize the efficient utilization of these convolutions.

\myPara{Analysis of image resolution.}
Here, we experiment to investigate the correlation between image resolution and the design of the multi-scale building block.
During inference, we apply test time augmentation performing multi-scaling transformations ($320 \times 320$, $640 \times 640$, and $1280 \times 1280$) on the image.
In addition, we also use these resolutions individually for evaluation. 
Note that the image resolution we used in training is $640\times640$.
The results are provided in \tabref{tab:different_resolution}.
Experimental results demonstrate a consistent trend: as the image resolution increases, $AP_s$ also increases. 
However, we can achieve higher $AP_l$ for low-resolution images.
This also verifies the effectiveness of our \protocolname{} protocol.

\subsection{Comparison with the State-of-the-Arts}

\myPara{Visualization comparisons.}
To evaluate which part of the image attracts the attention of the detector,
we use Grad-CAM \cite{selvaraju2017gradcam} to generate class response maps.
We visualize the class response maps generated from the neck of YOLOV7-T \cite{li2022yolov6}, RTMDet-T \cite{lyu2022rtmdet}, YOLOV8-N \cite{wang2022yolov7}, and \methodname{}-XS.
We also select typical images of different sizes, including small, medium, and large objects, from the MS COCO dataset \cite{lin2014microsoft}.
The visualization results are shown in \figref{fig:cam}.
YOLOV7-T, RTMDet-T, and YOLOV8-N fail to detect small, densely-packed objects, such as crowds of motorbikes and humans and ignore parts of them.
On the contrary, \methodname{}-XS exhibits a strong response for all objects in the class response maps, indicating its remarkable multi-scale feature representation ability. 
Furthermore, it highlights that our detector achieves excellent detection performance for objects of different sizes and images containing objects of various densities.

\myPara{Quantitative comparisons.}
In this section, we compare \methodname{} with the current state-of-the-art object detectors here and present the results in~\tabref{tab:comparison_sota}.
It is evident that \methodname{} achieves a remarkable speed-accuracy trade-off.
Compared with the second-best tiny detector, \ie RTMDet \cite{lyu2022rtmdet}, \methodname{}-XS reaches 42.8\% AP, which is 1.8\% AP higher than it with the ImageNet \cite{deng2009imagenet} pre-trained model.
\methodname{}-S achieves 45.4\% AP, which brings 0.8\% AP improvement with a faster inference speed compared with RTMDet. 
Furthermore, \methodname{} has a detection performance of 49.7\% AP, superior to the baseline with similar parameters and computational complexity.
In conclusion, 
\methodname{} proves to be a promising baseline for real-time object detection, 
offering powerful multi-scale feature representations.

\begin{table}[t]
    \centering
    \footnotesize
    \setlength{\tabcolsep}{5.2pt}
    \caption{
        Applications to other YOLOs.
        ``$^\dagger$'' refers to training with pertained models.
        All models are benchmarked under the same computational environment.
        \revised{The inference of YOLOv10 is with one-to-one training.}
    }
    \vspace{-10pt}
    \renewcommand\arraystretch{1.0}
    \begin{tabular}{lcccccc}
        Model                                    & \AP{}      &\AP{s}  &\AP{l} & Latency   & Params  & MACs    \\ \midrule[0.8pt]
        RTMDet-T$^\dagger$ \cite{lyu2022rtmdet}  & 41.0       & 20.7   & 58.0  & 6.5ms     & 4.9M    & 8.1G    \\ 
        \methodname{}-XS                         & 42.8       & 23.1   & 60.1  & 7.1ms     & 5.1M    & 8.7G    \\ 
        RTMDet-S$^\dagger$ \cite{lyu2022rtmdet}  & 44.6       & 24.2   & 61.8  & 7.3ms     & 8.9M    & 14.8G   \\ 
        \methodname{}-S                          & 45.4       & 25.9   & 62.4  & 7.3ms     & 8.7M    & 15.0G   \\ 
        RTMDet-M \cite{lyu2022rtmdet}            & 49.3       & 30.5   & 66.1  & 10.0ms    & 24.7M   & 39.3G   \\ 
        \methodname{}                            & 49.7       & 32.8   & 65.6  & 10.5ms    & 23.3M   & 38.8G   \\ \midrule[0.8pt] 
        YOLOv8-N \cite{jocher2023yolov8}         & 37.2       & 18.9   & 52.5  & 5.3ms     & 3.2M    & 4.4G    \\
        YOLOv8-MS-N                              & 40.2       & 20.9   & 55.5  & 6.5ms     & 2.9M    & 4.4G    \\ 
        YOLOv8-S \cite{jocher2023yolov8}         & 43.9       & 25.3   & 59.5  & 5.8ms     & 11.2M   & 14.4G   \\
        YOLOv8-MS-S                              & 46.2       & 27.0   & 62.7  & 6.9ms     & 9.5M    & 13.3G   \\  
        YOLOv8-M \cite{jocher2023yolov8}         & 49.8       & 32.6   & 65.9  & 8.3ms     & 25.9M   & 39.6G   \\ 
        YOLOv8-MS-M                              & 50.6       & 33.6   & 65.7  & 9.3ms     & 25.9M   & 35.2G   \\  \midrule[0.8pt] 
        \revised{YOLOv9-T \cite{wang2024yolov9}}          & 37.2       & 18.4   & 52.9  & 10.3ms    & 2.1M    & 4.1G    \\
        \revised{YOLOv9-MS-T}                             & 38.5       & 19.3   & 52.8  & 6.1ms     & 2.0M    & 4.2G    \\  \midrule[0.8pt] 
        \revised{YOLOv10-S \cite{wang2024yolov10}}         & 46.2       & 26.8   & 63.6  & 6.9ms     & 8.1M    & 12.4G   \\
        \revised{YOLOv10-MS-S}                             & 46.8       & 27.7   & 62.5  & 7.7ms     & 7.1M    & 11.5G   \\  
    \end{tabular}
    \label{tab:apply_other_yolos}
    \vspace{-5pt}
\end{table}

\myPara{Applications to other YOLOs.}
Our proposed methods can serve as a plug-and-play module for other YOLO models. 
To demonstrate the generalization ability of our method, 
we apply the proposed method to \revised{RTMDet, YOLOv8 \cite{jocher2023yolov8}, YOLOv9 \cite{wang2024yolov9} and YOLOv10 \cite{wang2024yolov10}}.
The results on MS COCO are listed in \tabref{tab:apply_other_yolos}.
With our methods, 
the AP scores of all scale baselines can be increased, 
with even fewer parameters and MACs. 
To be specific, our method advances the AP of YOLOv8-N, YOLOv8-S, and YOLOv8-M from 37.2\%, 43.9\%, and 49.8\% to 40.2\%, 46.2\%, and 50.6\%, respectively.
Additionally, our method improves the AP of objects at different scales, indicating its effectiveness in enhancing multi-scale ability.

\subsection{Application in other Tasks}

In this subsection, we extend our \methodname{} to three typical sub-tasks of object detection: instance segmentation, arbitrary-oriented object detection, and crowded scene object detection.
The datasets we use are COCO \cite{lin2014microsoft}, DOTA-v1.0 \cite{xia2018dota}, and CrowdHuman \cite{shao2018crowdhuman}, respectively.
The results are presented in \tabref{tab:instance_segmentation}, \tabref{tab:rotate_object_detection}, and \tabref{tab:crowd_detection}.
As illustrated in the results, our \methodname{} outperforms the strong baseline, demonstrating the robustness of our \methodname{} in different application environments.

\myPara{Instance segmentation.}
Instance segmentation is an extension task of object detection, aiming to delineate each instance in an image at the pixel level.
We apply our \methodname{} in the instance segmentation task on MS COCO \cite{lin2014microsoft}, as shown in \tabref{tab:instance_segmentation}.
Under the same training setting, 
\methodname{} achieves significant improvements, 
outperforming the baseline.
Specifically, the segmentation AP of our \methodname{} increases from 40.5\% to 42.8\% with an improvement of +2.3\%.

\begin{table}[ht]
  \vspace{-5pt}   
  \centering
  \footnotesize
  \caption{
    Quantitative results of YOLO-MS for the instance segmentation task. 
    The results are reported on the validation set of MS-COCO \cite{lin2014microsoft}.
    ``(LB)'' refers LetterBox resize proposed in \cite{jocher2020yolov5}.
    $^\dagger$ refers to models with the pre-trained model.
    The proposed \methodname{} results are marked in \textbf{gray}.
    The best results are in \textbf{bold}. 
  }
  \vspace{-10pt}
  \setlength{\tabcolsep}{2.5pt}
  \begin{tabular}{lccccccc}
    Model              &  Input size     & bbox \AP{}    & seg \AP{}     & seg \AP{s}    & Params & MACs  \\ \midrule[0.8pt]
    YOLOv5-N           &  640(LB)        & 27.6          &  23.4         & -             & 2.0M   & 3.6G  \\ 
    YOLOv5-S           &  640(LB)        & 37.6          &  31.7         & -             & 7.6M   & 13.2G \\ \midrule[0.8pt]
    RTMDet-T$^\dagger$ & $640\times640$ & 40.5          & 35.4          & 13.1         & 5.6M   & 11.8G \\ \rowcolor[gray]{.95}
    \methodname{}-XS   &  $640\times640$ & \textbf{42.3} & \textbf{36.6} & \textbf{15.6}  & 5.1M   & 12.9G  \\  
  \end{tabular}
  \label{tab:instance_segmentation}
  \vspace{-5pt}
\end{table}

\myPara{Arbitrary-oriented object detection.}
Arbitrary-oriented object detection is designed to detect objects in arbitrary orientations.
We compare \methodname{} with the baseline on the validation set of DOTA-v1.0 \cite{xia2018dota}, as presented in \tabref{tab:rotate_object_detection}.
Following the same training schedule as RTMDet-R, our \methodname{-R} outperforms the baseline at the tiny scale and particularly achieves the same performance as the small scale. 
Additionally, at other scales, our \methodname{-R} also introduces significant improvements over the baseline.
Given that objects in remote sensing scenarios are typically small, the advancements in DOTA v1.0 of 
RTMDet-R indicates that our \methodname{-R} enhances multi-scale capabilities.

\begin{table}[htp!]
    \vspace{-5pt}
    \centering
    \footnotesize
    \caption{
        Quantitative results of YOLO-MS for arbitrary-oriented object detection.
        The results are reported on the validation set of DOTA-v1.0 \cite{xia2018dota}.
        The proposed \methodname{} results are marked in \textbf{gray}.
        The best results are in \textbf{bold}. 
        \revised{
        The input size for evaluation and MACs computation is $1024\times1024$.
        }
    }
    \vspace{-10pt}
    \setlength{\tabcolsep}{8pt}
    \begin{tabular}{lcccccc}
            Model                          & \AP{}          & \AP{50}        & \AP{75}         & Params & MACs   \\ \midrule[0.8pt]
            RTMDet-R-T                     & 58.3           & 85.2           & 66.1            & 4.9M   & 21G    \\
            RTMDet-R-S                     & 62.0           & 88.1           & 70.6            & 8.9M   & 39G    \\ 
            RTMDet-R-M                     & 64.4           & 88.9           & 74.9            & 24.7M  & 100G   \\
            RTMDet-R-L                     & 66.3           & 89.4           & 76.9            & 52.3M  & 205G   \\ \rowcolor[gray]{.95}
            \methodname{}-R-XS             & 61.8           & 88.0           & 70.3            & 4.4M   & 22G    \\ \rowcolor[gray]{.95}
            \methodname{}-R-S              & 63.8           & 88.7           & 73.6            & 7.4M   & 38G    \\ \rowcolor[gray]{.95}
            \methodname{}-R                & 66.9           & 89.9           & 77.8            & 20.0M  & 99G    \\ \rowcolor[gray]{.95}
            \methodname{}-R-L              & \textbf{68.6}  & \textbf{90.6}  & \textbf{80.7}   & 42.7M  & 206G   \\ 
    \end{tabular}
    \label{tab:rotate_object_detection}
    \vspace{-5pt}
\end{table}

\myPara{Object detection for crowd scenes.}
Detecting objects in crowded scenarios is also a crucial topic in computer vision.
We evaluate the performance of our \methodname{} on the CrowdHuman \cite{shao2018crowdhuman} dataset, as shown in \tabref{tab:crowd_detection}. 
Consistent with prior works, we use the Average Precision (AP), mMR, and JT for evaluation.
mMR represents the average log miss rate over false positives per image ranging from $10^{-2}$ to 1.
It is sensitive to the False Positive Rate.
A lower mMR indicates better performance.
On the other hand, JI is the Jaccard Index, assessing the overlap between predictions and GTs. 
This metric is commonly employed to gauge the counting ability of a detector in crowded object detection scenarios. 
A higher JI signifies better performance.
Following the same training schedule as the baseline, our \methodname{} achieves a notable improvement with an AP increase of 1.2\%. 
Given the high-density and challenging nature of objects in the CrowdHuman dataset, these improvements underscore that our \methodname{} performs well in handling crowd scenes.

\begin{table}[ht]
    \vspace{-5pt}
    \centering
    \footnotesize
    \caption{
        Quantitative results of YOLO-MS for crowd scenes. 
        The results are reported on the validation set of CrowdHuman \cite{shao2018crowdhuman}.
        $^\dagger$ refers to models with pre-trained models.
        The proposed \methodname{} results are marked in \textbf{gray}.
        The best results are in \textbf{bold}. 
        \revised{
        The input size for evaluation and MACs computation is $640\times640$.
        }
    }
    \setlength{\tabcolsep}{8pt}
    \vspace{-10pt}
    \begin{tabular}{lccccc}
            Model              & \AP{}  & mMR \down  & JI    & Params &  MACs   \\ \midrule[0.8pt]
            RTMDet-T$^\dagger$  & 85.8   & 47.2       & 76.0  & 4.9M   &  8.0G   \\ \rowcolor[gray]{.95}
            \methodname{}-XS   & \textbf{87.0}   & \textbf{46.3}       & \textbf{78.1}  & 5.1M   &  8.6G   \\ 
    \end{tabular}
    \label{tab:crowd_detection}
    \vspace{-5pt}
\end{table}

\myPara{Object detection for underwater scenes}.
\revised{
RUDO \cite{fu2023rethinking} dataset contains 14,000 high-resolution underwater images, 74,903 labeled objects, and 10 common aquatic categories.
As shown in~\tabref{tab:rudo}, our \methodname{} achieves the \sota{} performance and outperforms the baseline, which is trained under the same schedule, with an AP increase of 0.5\%. 
Given the challenging underwater scenes in the RUDO dataset, these improvements underscore that our \methodname{} performs well in handling different conditions.
}

\begin{table}[htp!]
    \vspace{-5pt}
    \centering
    \footnotesize
    \renewcommand\arraystretch{1}
    \renewcommand{\tabcolsep}{6pt}
    \caption{
        \revised{
        Quantitative results of YOLO-MS for underwater scenes.
        The results are reported on the validation set of RUDO \cite{fu2023rethinking}.
        $^\dagger$ refers to models with pre-trained models.
        The proposed \methodname{} results are marked in \textbf{gray}.
        The best results are in \textbf{bold}.
        The input size for MACs computation is $640\times640$.
        }
    }
    \label{tab:rudo}
    \vspace{-10pt}
    \begin{tabular}{lcccc}
            Model                                   & Input size      & \AP{}           & Params &  MACs   \\  \midrule[0.8pt]
            Cascade RCNN-R50 \cite{cai2018cascade}  & $1333\times800$ & 55.3            & 77.3M  &  1709G  \\
            ATSS-R50 \cite{zhang2020bridging}       & $1333\times800$ & 55.7            & 32.3M  &  82.0G  \\
            TOOD-R50 \cite{feng2021tood}            & $1333\times800$ & 57.4            & 32.2M  &  80.4G  \\
            RTMDet-T$^\dagger$                      & $640\times640$  & 59.2            & 4.9M   &  8.1G   \\ \rowcolor[gray]{.95}
            \methodname{}-XS                        & $640\times640$  & \textbf{59.7}   & 5.1M   &  8.6G   \\ 
    \end{tabular}
    \vspace{-5pt}
\end{table}

\myPara{Object detection for foggy scenes}.
\revised{
RTTS \cite{li2018benchmarking} dataset consists of 4,322 foggy images with 5 categories: bicycle, bus, car, motorbike, and person.
As shown in~\tabref{tab:rtts}, our \methodname{} achieves the \sota{} performance and outperforms the baseline, which is trained under the same schedule, with an AP increase of 0.3\%. 
Given the challenging hazing scenes in the RTTS dataset, these improvements underscore that our \methodname{} performs well in handling different weather conditions.
}

\vspace{-5pt}
\begin{table}[htp!]
    \vspace{-5pt}
    \centering
    \footnotesize
    \renewcommand\arraystretch{1}
    \renewcommand{\tabcolsep}{5pt}
    \caption{
        \revised{
        Quantitative results of YOLO-MS for foggy scenes.
        The results are reported on the validation set of RTTS \cite{li2018benchmarking}.
        $^\dagger$ refers to models with pre-trained models.
        The proposed \methodname{} results are marked in \textbf{gray}.
        The best results are in \textbf{bold}.
        The input size for MACs computation is $640\times640$.
        }
    }
    \label{tab:rtts}
    \vspace{-10pt}
    \begin{tabular}{lcccc}
            Model                                  &   Input size       & \AP{}          & Params &  MACs   \\  \midrule[0.8pt]
            Cascade RCNN-R50 \cite{cai2018cascade} &   $1333\times800$  & 50.8           & 77.3M  &  1709G   \\
            ATSS-R50 \cite{zhang2020bridging}      &   $1333\times800$  & 48.2           & 32.3M  &  82.0G  \\
            TOOD-R50 \cite{feng2021tood}           &   $1333\times800$  & 50.8           & 32.2M  &  80.4G  \\
            RTMDet-T$^\dagger$                     &   $640\times640$   & 63.6           & 4.9M   &  8.1G   \\ \rowcolor[gray]{.95}
            \methodname{}-XS                       &   $640\times640$   & \textbf{63.9}  & 5.1M   &  8.6G   \\ 
    \end{tabular}
    \vspace{-5pt}   
\end{table}

\begin{table*}[t]
    \vspace{-5pt}
    \centering
    \footnotesize
    \setlength{\tabcolsep}{15pt}
    \caption{
    \revised{
    Quantitative results of YOLO-MS for RAW image in diverse conditions.
    The results are reported on the validation set of AODRaw \cite{li2024aodraw}.
    $^\dagger$ refers to models with pre-trained models.
    The proposed \methodname{} results are marked in \textbf{gray}.
    The best results are in \textbf{bold}.
    The input size for evaluation is $1280\times1280$, while for the MACs computation is $640\times640$.
     }
    }
    \vspace{-10pt}
    \begin{tabular}{lccclllcc}
    Model                                    & \AP{}  & \AP{50}& \AP{75} & \AP{s}& \AP{m}&\AP{l}& Params &  MACs   \\ \midrule
    YOLOX-T \cite{ge2021yolox}               & 16.4   & 32.1   &  14.9   & 6.8   & 23.2  & 29.4 &  5.1M  &  7.6G   \\
    YOLOv6-N \cite{li2022yolov6}             & 18.0   & 30.0   &  18.0   & 7.6   & 24.4  & 32.8 &  4.3M  &  5.5G   \\
    YOLOv8-N \cite{jocher2023yolov8}         & 18.9   & 32.0   &  18.8   & 8.9   & 26.5  & 33.2 &  3.0M  &  4.4G   \\
    RTMDet-T$^\dagger$ \cite{lyu2022rtmdet}  & 24.3   & 40.0   &  24.7   & 11.5  & 34.0  & 40.6 &  4.9M  &  8.1G   \\ \rowcolor[gray]{.95}
    \methodname{}-XS                         & \textbf{25.4}   & \textbf{40.6}   &  \textbf{26.1}   & \textbf{11.8}  & \textbf{34.4}  & \textbf{42.7} &  5.1M  &  8.7G   \\ 
    \end{tabular}
    \label{tab:aodraw}
    \vspace{-10pt}
\end{table*}

\myPara{Object detection for RAW image in diverse conditions.} 
\revised{
AODRaw \cite{li2024aodraw} dataset includes 7,785 high-resolution real RAW images with 135,601 annotated instances spanning 62 categories. 
It captures a wide range of indoor and outdoor scenes under nine distinct light and weather conditions.
As shown in~\tabref{tab:rtts}, our \methodname{} achieves the \sota{} performance and outperforms the baseline, which is trained under the same schedule, with a notable AP increase of 1.1\%. 
Given the various scenes and RAW image input in the AODRaw dataset, these improvements underscore that our \methodname{} performs well in handling different indoor and outdoor weather conditions, not only for RGB images.
}

\section{Conclusions and Discussions}

This paper proposes a high-performant real-time object detector with a reasonable computational cost.
To accomplish this goal, we investigate the influence of feature distribution and convolution with varying kernel sizes and construct an encoder that can powerfully extract multi-scale feature representations.
Our experimental studies reveal that the proposed \blockname{} combined with \attentionname{} and \protocolname{} protocol significantly enhance the speed-accuracy trade-off of our detector, surpassing other real-time detectors.
We hope this work will bring new insights to the object detection community.


{\small
\bibliographystyle{plain}
\bibliography{egbib}
}

\newcommand{\addPhoto}[1]{\includegraphics[width=1in,height=1.15in,clip,keepaspectratio]{figures/bio_fig/#1}}


\vspace{-30pt}
\begin{IEEEbiography}[\addPhoto{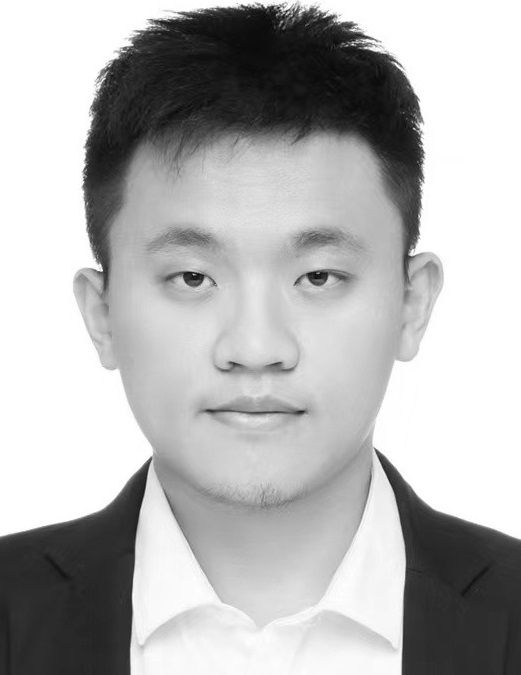}]
{Yuming Chen} received his B.E. degree in computer science from Lanzhou University in 2022. 
He is currently a Ph.D. candidate at Media Computing Lab, Nankai University, supervised by Prof. Ming-Ming Cheng and Prof. Qibin Hou. 
His research interests include vision understanding, object detection and knowledge distillation.
\end{IEEEbiography}

\vspace{-30pt}
\begin{IEEEbiography}[\addPhoto{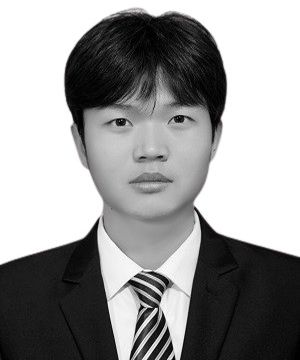}]
{Xinbin Yuan} received his B.E. degree in Information Management and Information Systems from Northwest A\&F University in 2023. 
He is a Master's student at the Media Computing Lab at Nankai University, supervised by Prof. Ming-Ming Cheng and Prof. Qibin Hou. 
His research interests include object detection and remote sensing.
\end{IEEEbiography}

\vspace{-30pt}
\begin{IEEEbiography}[\addPhoto{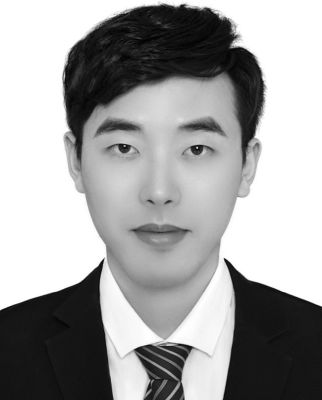}]
{Jiabao Wang} received his M.E. degree in automation from Northwestern Polytechnical University in 2022. 
He is currently a Ph.D. candidate at Media Computing Lab, Nankai University, supervised by Prof. Ming-Ming Cheng and Prof. Qibin Hou. 
His research interests cover various computer vision and machine learning topics, such as 2D/3D/oriented object detection and 3D perception in autonomous driving.
\end{IEEEbiography}

\vspace{-30pt}
\begin{IEEEbiography}[\addPhoto{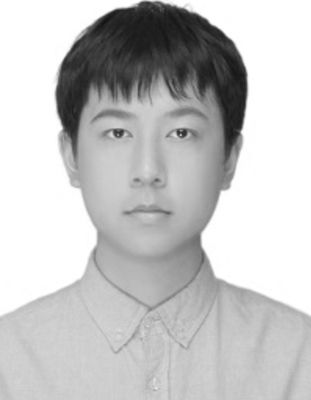}]
{Ruqi Wu} received his B.E. degree from the School of Computer Science and Artificial Intelligence, Wuhan University of Technology, in 2022. 
He is currently a Ph.D. candidate at the Media Computing Lab at Nankai University, supervised by Prof. Ming-Ming Cheng and Prof. Chun-Le Guo. 
His research interests include image/video generation and image processing.
\end{IEEEbiography}

\vspace{-30pt}
\begin{IEEEbiography}[\addPhoto{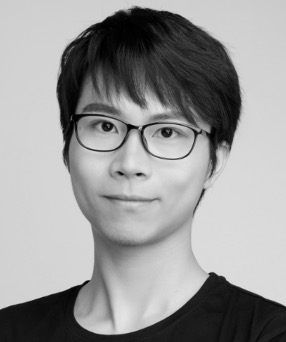}]{Xiang Li}
is an Associate Professor at the College of Computer Science, Nankai University.
He obtained his Ph.D. degree from Nanjing University of Science and Technology, Jiangsu, China, in 2020. 
His research interests include CNN/Transformer backbone, object detection,
knowledge distillation, and self-supervised learning. He has published 30+ papers in top journals and conferences such as TPAMI, CVPR,
NeurIPS, etc.
\end{IEEEbiography}

\vspace{-30pt}
\begin{IEEEbiography}[\addPhoto{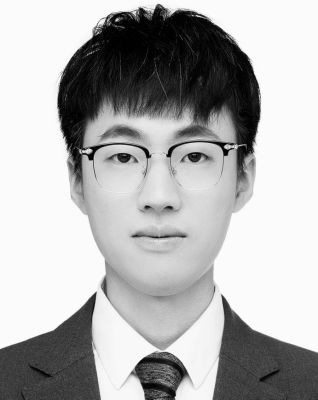}]
{Qibin Hou} received his Ph.D. degree from the School of Computer Science, Nankai University. Then, he worked at the National University of Singapore as a research fellow. 
Now, he is an associate professor at the School of Computer Science, Nankai University. He has published over 40 papers in top conferences/journals, including T-PAMI, CVPR, ICCV, NeurIPS, etc. His research interests include deep learning, image processing, and computer vision.
\end{IEEEbiography}

\vspace{-30pt}
\begin{IEEEbiography}[\addPhoto{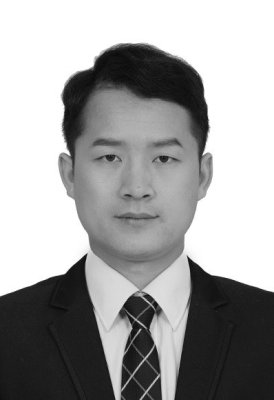}]{Ming-Ming Cheng} 
received his PhD degree from Tsinghua University in 2012.
Then, he did 2 years research fellow, with Prof. Philip Torr in Oxford.
He is now a professor at Nankai University, leading the Media Computing Lab.
His research interests include computer graphics, computer vision, and image processing. 
He received research awards, including the National Science Fund for Distinguished Young Scholars and the ACM China Rising Star Award.
He is on the editorial boards of IEEE TPAMI and IEEE TIP.
\end{IEEEbiography}

\vfill
\end{document}